\documentclass[a4paper,fleqn]{cas-sc}

\usepackage[authoryear,longnamesfirst]{natbib}
\usepackage{amsfonts}
\usepackage{graphicx}
\usepackage{subfigure}
\usepackage{mathrsfs}
\usepackage{amsmath}
\usepackage{amssymb}
\usepackage{float}
\usepackage{bm}
\usepackage{booktabs}
\usepackage{threeparttable}
\usepackage{multirow }
\usepackage[ruled, linesnumbered]{algorithm2e}
\usepackage{cite}
\usepackage{enumitem}
\usepackage{hyperref}

\newtheorem{thm}{Theorem}

\newtheorem{lem}{Lemma}
\newtheorem{Pro}{Proposition}

\newproof{pf}{Proof}

\def\0{{\bf 0}}
\def\T{{\bf T}}
\def\cT{\mathcal{T}}

\def\bga{{\bm \gamma}}
\def\bbR{\mathbb{R}}
\def\cV{\mathcal{V}}
\def\be{\bm{\beta}}
\def\cS{\mathcal{S}}
\def\bSigma{{\bf\Sigma}}
\def\bxi{{\bm \xi}}
\def\1{{\bf 1}}
\def\St{{\rm St}}

\def\tr{\mathrm {tr}}

\def\tr{\mathrm {tr}}
\def\U{{\bf U}}
\def\V{{\bf V}}
\def\A{{\bf A}}

\def\B{{\bf B}}
\def\D{{\bf D}}

\def\C{{\bf C}}
\def\S{{\bf S}}
\def\Q{{\bf Q}}
\def\L{{\bf L}}
\def\I{{\bf I}}
\def\E{{\bf E}}

\def\X{{\bf X}}
\def\Z{{\bf Z}}

\def\Y{{\bf Y}}

\def\tr{\mathrm {tr}}

\def\diag{\mathrm {diag}}

\def\H{\bf H}

\def\tr{\mathrm {tr}}

\newcommand{\independent}{\;\, \rule[0em]{.03em}{.67em} \hspace{-.25em}
	\rule[0em]{.65em}{.03em} \hspace{-.25em}
	\rule[0em]{.03em}{.67em}\;\,}

\begin{document}
\let\WriteBookmarks\relax
\def\floatpagepagefraction{1}
\def\textpagefraction{.001}

\shorttitle{R. Wu and X. Chen/Preprint}
\shortauthors{R. Wu and X. Chen.}

\title[mode = title]{MM Algorithms for Distance Covariance based Sufficient Dimension Reduction and Sufficient Variable Selection}

\author{Runxiong Wu}
\ead{11930643@mail.sustech.edu.cn}

\author{Xin Chen}
\ead{chenx8@sustech.edu.cn}
\cormark[1]
\cortext[cor1]{Corresponding author}

\address{Department of Statistics \& Data Science, Southern University of Science and Technology, Shenzhen 518055, China}

\begin{abstract}
Sufficient dimension reduction (SDR) using distance covariance (DCOV) was recently proposed as an approach to dimension-reduction problems. Compared with other SDR methods, it is model-free without estimating link function and does not require any particular distributions on predictors. However, the DCOV-based SDR method involves optimizing a nonsmooth and nonconvex objective function over the Stiefel manifold. To tackle the numerical challenge, the original objective function is equivalently formulated into a DC (Difference of Convex functions) program and an iterative algorithm based on the majorization-minimization (MM) principle is constructed. 
At each step of the MM algorithm, one iteration of Riemannian Newton's method is taken to solve the quadratic subproblem on the Stiefel manifold inexactly. In addition, the algorithm can also be readily extended to sufficient variable selection (SVS) using distance covariance. Finally, the convergence property of the proposed algorithm under some regularity conditions is established. Simulation and real data analysis show our algorithm drastically improves the computation efficiency and is robust across various settings compared with the existing method. \verb|Matlab| codes implementing our methods and scripts for regenerating the numerical results are available at \url{https://github.com/runxiong-wu/MMRN}.
\end{abstract}


\maketitle

\section{Introduction}
In regression analysis, sufficient dimension reduction (SDR) provides a useful statistical framework to analyze a high-dimensional dataset without losing any information. It finds the fewest linear combinations of predictors that capture a full regression relationship. Let $Y$ be an univariate response and $X=(x_1,\ldots,x_p)^{\top}$ be a $p\times 1$ predictor vector, SDR aims to find a $p\times d$ matrix $\be$ such that
\begin{equation}\label{eqn1.1}
Y \independent X | \be^{\top} X, \tag{1.1}
\end{equation}
where $\independent$ denotes the statistical independence. The column space of $\be$ satisfying \eqref{eqn1.1} is called a dimension reduction subspace. Under mild conditions \citep{cook1996graphics,yin2008successive}, the intersection of all the dimension reduction subspaces exists and is unique. In this case, if the intersection itself is also a dimension reduction subspace, we call it the central subspace \citep{cook1994interpretation,cook1996graphics} for the regression of $Y$ on $X$ and denote it by $\mathcal{S}_{Y|X}$. Note that the dimension of $\mathcal{S}_{Y|X}$ denoted by $\mbox{dim}(\mathcal{S}_{Y|X})$ is usually much smaller than the original predictor's dimension $p$. Thus, we reduce the dimensionality of the predictor space. The primary interest of SDR is to find such central subspace $\mathcal{S}_{Y|X}$.

Since the introduction of sliced inverse regression \citep[SIR;][]{li1991sliced} and sliced average variance estimation \citep[SAVE;][]{cook1991sliced}, many methods have been proposed for estimating the basis of $\cS_{Y|X}$, including  inverse regression \citep[IR;][]{cook2005sufficient}, directional regression \citep[DR;][]{li2007directional}, minimum average variance estimation method \citep[MAVE;][]{xia2002adaptive}, sliced regression \citep[SR;][]{wang2008sliced}, ensemble approach \citep{yin2011sufficient}, Fouriers transform approach \citep{zhu2006fourier}, integral transform method \citep{zeng2010integral}, Kullback-Leibler distance based estimator \citep{yin2005direction}, likelihood based method \citep{cook2009likelihood}, and semiparametric approach \citep{ma2012semiparametric}, etc.

All of the aforementioned dimension reduction methods require certain conditions on the predictors or complicated smoothing technique. In reality, these conditions are not easy to be verified and the results of these methods may be misleading if the conditions are violated. Recently, \citet{sheng2013direction,sheng2016sufficient} proposed a method using distance covariance \citep[DCOV;][]{szekely2007measuring,szekely2009brownian} for estimating the central subspace $\cS_{Y|X}$. Distance covariance is an elegant measure that quantifies the dependence strength between two random vectors. Consequently, the DCOV-based SDR method requires only mild conditions on the predictors and does not require any link function or nonparametric estimation. It can be also easily extended to handle regression with multivariate responses.

The most challenging part of the DCOV-based SDR methods is that it involves solving a nonconvex and nonsmooth optimization problem over the Stiefel manifold. The present work \citep[e.g.,][]{sheng2013direction,sheng2016sufficient,chen2018efficient} tackled the problems by using sequential quadratic programming \citep[SQP;][chap. 6]{gill1981practical}. The SQP method works well when the dimension $p$ and the sample $n$ is not too large, but optimization is often computationally difficult for moderately high dimensional settings. Another method that seems to work is to use the Matlab package \verb|manopt| by \citet{boumal2014manopt}. This package provides iterative Riemannian optimization techniques, including Trust-regions, BFGS, SGD, Nelder-Mead, and so on. Unfortunately, directly applying this package to solve the DCOV-based SDR problems may often crash since it needs the analytical first-order derivative function. Beyond above, the literature on solving this kind of problem is scarce.

In this article, we propose a new algorithm which presents three major contributions to the literature of sufficient dimension reduction and manifold optimization. First, we novelly write the DCOV objective function of the model as a difference of convex functions equivalently. Therefore we design a highly efficient algorithm for solving the corresponding optimization problem based on the new objective function form. Second, we construct the convergence property of the proposed algorithm over the Stiefel manifold. Third, we extend our method to sufficient variable selection based on distance covariance. Simulation studies show our algorithm is ten to hundred times faster than the methods relying on SQP algorithm.

A toy example is given to visualize what SDR does and to see the performance of our algorithm and the competitor's. In this example, we generate 800 independent copies one time from
\begin{equation*}
X = {\bm\Gamma} [\cos(2\pi Y), \sin(2\pi Y)]^\top + 0.1{\bm\Phi}^{1/2}\epsilon, \nonumber
\end{equation*} where
\begin{equation*}
\bm\Gamma = \left(
\begin{array}{ccccccc}
1 & 1 & \ldots & 1 & 1\\
1 & -1 & \ldots & 1 & -1 \\
\end{array}
\right)^\top  \in \mathbb{R}^{20 \times 2},
\end{equation*} $Y$ is generated from uniform distribution over interval $(0, 1)$, $\Phi_{ij}=0.5^{|i-j|}$ and $\epsilon$ is a standard normal error. In the following figure, we can see how the first two SDR components recover a circle pattern. Our algorithm (MMRN, see details in a later chapter) is about 20 time faster than the competitor.
\begin{figure}[!htbp]
	\centering	
	\includegraphics[scale=.75]{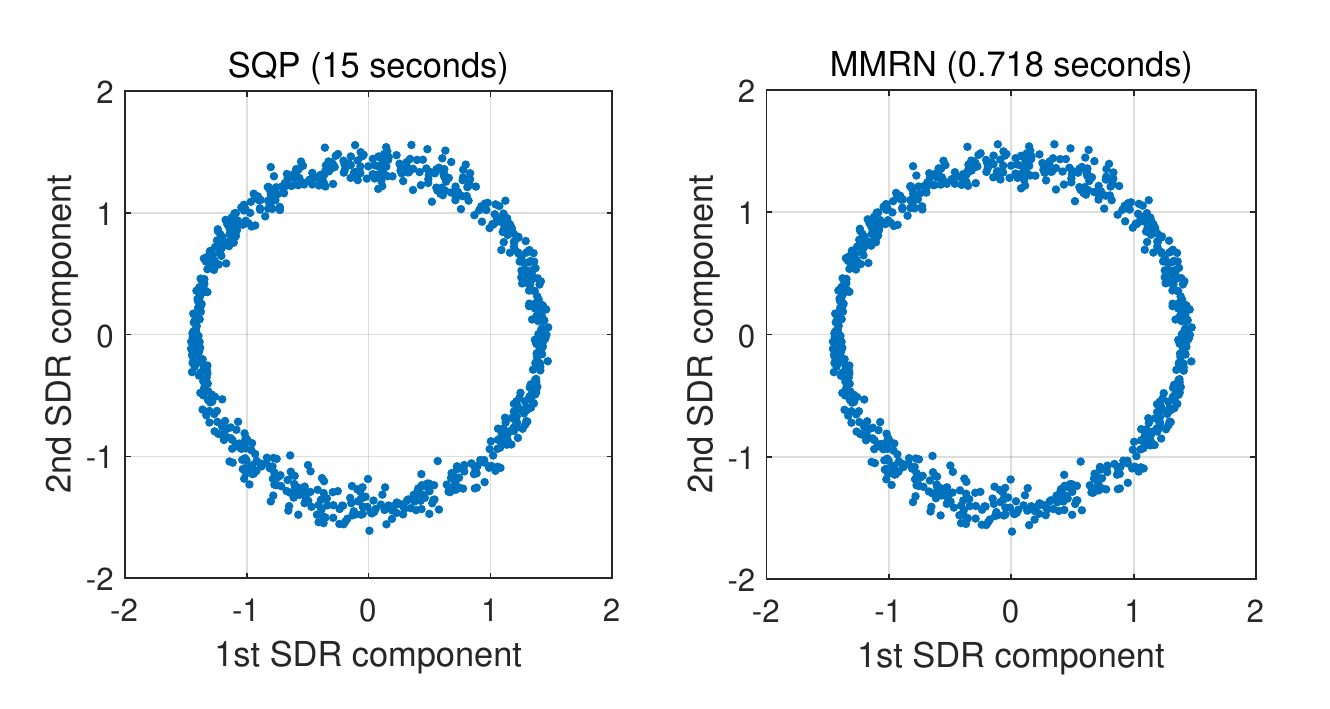}	
	\caption{Computational performance comparison.}
	\label{FIG:1}
\end{figure}

\subsection{Notation and the Stiefel Manifold}
The following notations and knowledge about the Stiefel manifold discussed in \citet{absil2009optimization,edelman1998geometry} will be used in our exposition. The trace of a matrix $\A$ is $\tr(\A)$ and the Euclidean inner product of two matrices $\A,\B$ is $\langle \A, \B \rangle=\tr(\A^{\top}\B)$. We use $\|\cdot\|_2$ and $\|\cdot\|_{\rm F}$ to denote the Euclidean norm of a vector and the Frobenius norm of a matrix respectively. The notation $\St(d,p)=\left\{ \bga\in\bbR^{p\times d}|\bga^{\top}\bga=\I_d \right\}$ with $d\leq p$ is referred to the Stiefel manifold and $\cT_{\bga}\St(d,p)$ is the tangent space to $\St(d,p)$ at a point $\bga\in\St(d,p)$. According to \citet{edelman1998geometry}, $\cT_{\bga}\St(d,p)=\left\{\bga\U+\bga_{\perp}\V|\U\in{\rm Skew}(d),\V\in\bbR^{(p-d)\times d} \right\}$. Here $\bga_{\perp}$ is the orthogonal complement of $\bga$ and ${\rm Skew}(d)$ denotes the set of $d\times d$ skew-symmetric matrices. We use ${\rm vec}({\bf W})$ to denote the vector formed by stacking the column vectors of ${\bf W}$. For a skew-symmetric matrix ${\bf W}\in {\rm Skew}(d)$, ${\rm veck}({\bf W})$ denotes a $d(d-1)/2$-dimensional column vector obtained by stacking the columns of the lower triangular part of ${\bf W}$. For a square matrix ${\bf W}$, we use ${\rm sym}({\bf W})=\left( {\bf W}+{\bf W}^{\top}\right)/2$ and ${\rm skew}({\bf W})=\left( {\bf W}-{\bf W}^{\top}\right)/2$ to denote the symmetric and skew-symmetric parts of ${\bf W}$ respectively. Induced from the Euclidean inner product, the Riemannian metric on $\St(d,p)$ we consider here is defined as $\langle \bxi_1,\,\bxi_2  \rangle_{\bga}=\tr(\bxi_1^{\top}\bxi_2),\; \mbox{for any } \bxi_1,\bxi_2 \in \cT_{\bga}\St(d,p)$. Under this metric, the orthogonal projection of ${\bf W}$ onto the tangent space $\cT_{\bga}\St(d,p)$ is expressed as $\mbox{Proj}_{\cT_{\bga}\St(d,p)}({\bf W})={\bf W}-\bga {\rm sym} \left(\bga^{\top}{\bf W}\right)$. Let $f$ be a smooth function and $\nabla f$ be the Euclidean gradient, the Riemannian gradient of point $\bga \in \St(d,p)$ is defined as ${\rm grad}f(\bga)= \mbox{Proj}_{\cT_{\bga}\St(d,p)}(\nabla f(\bga))$. Correspondingly, the Riemannian Hessian of point $\bga \in \St(d,p)$ acting on $\bxi \in \cT_{\bga}\St(d,p)$ is defined as $\mbox{Hess}\,f(\bga)[\bxi] =\mbox{Proj}_{\cT_{\bga}\St(d,p)}\left({\rm D}(\mbox{grad}\,f)(\bga)[\bxi]\right)$ and ${\rm D}(\mbox{grad}\,f)(\bga)[\bxi]$ is the directional derivative of $\mbox{grad}\,f(\bga)$ along the direction $\bxi$. We use Retr to denote the retraction operation. For the Stiefel manifold, the QR retraction is used in the article.

\subsection{Organization}
The rest of the article is organized as follows. Section 2 reviews briefly key knowledge of the DCOV-based SDR method and illustrates our motivation. Section 3 describes the proposed algorithm for solving DCOV-based SDR models in details and Section 4 extends the proposed algorithm to DCOV-based SVS models. In Section 5, we evaluate the superior numeric performance of the proposed algorithm through various simulation studies. Finally, we draw some concluding remarks about the article in Section 6. All proofs are given in the Appendix.

\section{Background Review and Motivation}
\subsection{DCOV-based SDR Model}
Let $(\X,\Y)=\left\{ (X_i,Y_i): i=1,\ldots,n \right\}$ be a random sample from $(X,Y)$. $\X$ denotes a $p\times n$ data matrix and $\Y$ denotes a $1 \times n$ response data matrix. We present here an univariate response, however, the method can naturally be extended to multivariate responses without any issue due to the nature of DCOV. The empirical solution of DCOV-based SDR method for these $n$ observations relies on solving the following objective function:
\begin{alignat}{1}\label{eqn2.1}
\underset{ \be\in\bbR^{p\times d} }{\mbox{max}} \; & \cV_{n}^2(\be^{\top}\X,\Y) := \frac{1}{n^2} \sum_{k,l=1}^{n} A_{kl}({\be})B_{kl},  \mbox{ s.t. } \be^{\top} \widehat{\bSigma}_{X} \be =\I_d, \tag{2.1}
\end{alignat}
where $\widehat{\bSigma}_X$ is the sample covariance matrix of $X$, $\I_d$ is a $d$-dimensional identity matrix and for $k,l=1,\ldots,n,$
\begin{eqnarray*}
	A_{kl}({\be}) &=& a_{kl}({\bm\beta})- \overline{a}_{k\cdot }({\bm\beta})-  \overline{a}_{\cdot l}({\bm\beta})+\overline{a}_{\cdot \cdot}({\bm\beta}), \\
	a_{kl}({\bm\beta}) &=& \| \bm{\beta}^{\top} X_k-\bm{\beta}^{\top} X_l \|_2,
	\;  \overline{a}_{k\cdot }(\bm{\beta})=\frac{1}{n} \sum_{l=1}^{n} a_{kl}(\bm{\beta}),   \\
	\overline{a}_{\cdot l}({\bm\beta})  &=&  \frac{1}{n} \sum_{k=1}^{n} a_{kl}({\bm\beta}),\; \overline{a}_{\cdot \cdot}({\bm\beta})  = \frac{1}{n^2} \sum_{k,l=1}^{n} a_{kl}({\bm\beta}).
\end{eqnarray*}
Similarly, define $b_{kl}=\|Y_k-Y_l\|_2$ and $B_{kl}= b_{kl}- \overline{b}_{k\cdot }-  \overline{b}_{\cdot l}+\overline{b}_{\cdot \cdot}$. \citet{sheng2013direction,sheng2016sufficient} showed that under mild conditions, the solution of the above problem $(\ref{eqn2.1})$ is a $\sqrt{n}$-consistent estimator of a basis of $\cS_{Y|X}$.

\subsection{Motivation}
In the Appendix of \citet*{szekely2007measuring}, it was proved that $\cV_{n}^2(\be^{\top}\X,\Y)$ has another expression, i.e.,
\begin{equation}\tag{2.2}\label{eqn2.2}
\cV_{n}^2(\be^{\top}\X,\Y)=S_1+S_2-2S_3,
\end{equation}
where
\begin{equation} \tag{2.3}\label{eqn2.3}
\begin{aligned}
S_1 &= \frac{1}{n^2}\sum_{k,l=1}^{n} a_{kl}(\be)b_{kl},\\
S_2 &= \frac{1}{n^2}\sum_{k,l=1}^{n} a_{kl}(\be)\frac{1}{n^2}\sum_{k,l=1}^{n} b_{kl}=\frac{1}{n^2}\sum_{k,l=1}^{n} a_{kl}(\be)\overline{b}_{\cdot\cdot},\\
S_3 &= \frac{1}{n^3}\sum_{k=1}^{n}\sum_{l,m=1}^{n}a_{kl}(\be)b_{km}= \frac{1}{n^2}\sum_{k,l=1}^{n} a_{kl}(\be)\overline{b}_{k\cdot}.
\end{aligned}
\end{equation}
Notice that $\displaystyle \frac{1}{n^2}\sum_{k,l=1}^{n} a_{kl}(\be)\overline{b}_{k\cdot}= \frac{1}{n^2}\sum_{k,l=1}^{n} a_{kl}(\be)\overline{b}_{\cdot l}$ because for any $k,l=1,\ldots,n$, $a_{kl}(\be)\overline{b}_{k\cdot}=a_{lk}(\be)\overline{b}_{\cdot k}$. Then, we have the following way to express $2S_3$:
\begin{equation}\tag{2.4}\label{eqn2.4}
2S_3=\frac{1}{n^2}\sum_{k,l=1}^{n} a_{kl}(\be)\overline{b}_{k\cdot}+ \frac{1}{n^2}\sum_{k,l=1}^{n} a_{kl}(\be)\overline{b}_{\cdot l}=\frac{1}{n^2}\sum_{k,l=1}^{n} a_{kl}(\be) \left( \overline{b}_{k\cdot}+\overline{b}_{\cdot l}\right).
\end{equation}
Substituting equations $(\ref{eqn2.3})$ and $(\ref{eqn2.4})$ into $(\ref{eqn2.2})$, we obtain
\begin{equation}\tag{2.5}\label{eqn2.5}
\begin{aligned}
\cV_{n}^2(\be^{\top}\X,\Y) &= \frac{1}{n^2}\sum_{k,l=1}^{n} a_{kl}(\be)b_{kl}+\frac{1}{n^2}\sum_{k,l=1}^{n} a_{kl}(\be)\overline{b}_{\cdot\cdot}-\frac{1}{n^2}\sum_{k,l=1}^{n} a_{kl}(\be) \left( \overline{b}_{k\cdot}+\overline{b}_{\cdot l}\right), \\
&= \frac{1}{n^2}\sum_{k,l=1}^{n} a_{kl}(\be) \left( b_{kl}+ \overline{b}_{\cdot\cdot} -\overline{b}_{k\cdot}-\overline{b}_{\cdot l} \right),\\
&= \frac{1}{n^2}\sum_{k,l=1}^{n} a_{kl}(\be) B_{kl}.
\end{aligned}
\end{equation}
In addition, it can be verified that $\sum_{k,l=1}^{n}B_{kl}=0$ and $a_{kl}(\be)$ is convex with respect to $\be$. These details make us notice that the objective function $(\ref{eqn2.5})$ can have a difference of convex functions decomposition (DC). Indeed, we can write the function $(\ref{eqn2.5})$ into a DC formulation
\begin{equation}\tag{2.6}\label{eqn2.6}
\cV_{n}^2(\be^{\top}\X,\Y)=\left(  \frac{1}{n^2}\sum_{k,l=1}^{n}a_{kl}(\be)B_{kl}I(B_{kl}>0)\right)- \left( -\frac{1}{n^2}\sum_{k,l=1}^{n}a_{kl}(\be)B_{kl}I(B_{kl}<0)\right)
\end{equation}
through the indicator function $I(\cdot)$. This equivalent function form $(\ref{eqn2.6})$ motivates us to design a highly efficient algorithm from the viewpoint of difference convex algorithm \citep[DCA;][]{tao1997convex}. More details about DCA and some of its recent developments
can be found in \citet{tao2005dc,le2018dc}; \citet{tao1997convex,tao1998dc,dinh2014recent}.

Thus, the objective function $(\ref{eqn2.1})$ of the DCOV-based SDR model can be equivalently transformed to
\begin{equation}\label{eqn2.7}\tag{2.7}
\begin{split}
\underset{ \be \in\bbR^{p\times d}}{\mbox{max}} \; & \cV_{n}^2(\be^{\top}\X,\Y) := \frac{1}{n^2} \sum_{k,l=1}^{n} a_{kl}({\be})B_{kl}, \mbox{ s.t. }\; \be^{\top} \widehat{\bSigma}_{X} \be =\I_d.
\end{split}
\end{equation}
Let $\bga={\widehat{\bSigma}_X}^{\frac 12}\be$ and $\Z={\widehat{\bSigma}_X}^{-\frac 12} \X$, the above function $(\ref{eqn2.7})$ can be rewritten as
\begin{equation}\label{eqn2.8}\tag{2.8}
\begin{split}
\underset{ \bga}{\mbox{max}}\; \cV_{n}^2(\bga^{\top}\Z,\Y):=\frac{1}{n^2} \sum_{k,l=1}^{n} a_{kl}({\bga})B_{kl}, \mbox{ s.t. } \; \bga \in \St(d,p),
\end{split}
\end{equation}
where $a_{kl}(\bga)=\|\bga^{\top}Z_k-\bga^{\top}Z_l\|_2$. In later sections, we will make full use of the equivalent form $(\ref{eqn2.8})$ rather than $(\ref{eqn2.7})$.

\section{Methodology}

\subsection{Preliminaries}
In fact, DCA is based on MM algorithm which is a principle of designing algorithms. The idea of designing a MM algorithm for finding $\hat{x}=\underset{ x \in \mathscr{X} }{\arg\max}\, f(x)$ where ${\mathscr X}$ is the constraint region is as follows. At each iterate $x^{(t)}$, we need to  construct a surrogate function $g(x|x^{(t)})$ satisfying
\begin{equation*}
\begin{split}
f(x^{(t)}) &= g(x^{(t)}|x^{(t)})  \\
f(x) &\geq g(x|x^{(t)}),\qquad \mbox{for any  $x \in \mathscr{X} $ }.
\end{split}
\end{equation*}
Then, MM algorithm updates the estimation with
\begin{equation*}
x^{(t+1)} =\underset{ x \in \mathscr{X}   }{\arg\max }\; g(x|x^{(t)}).
\end{equation*}
Because
\begin{equation*}
f(x^{(t+1)}) \geq g(x^{(t+1)}|x^{(t)}) \geq g(x^{(t)}|x^{(t)}) = f(x^{(t)}),
\end{equation*}
the iterate estimates generated by MM algorithm drive the objective function uphill. Under mild conditions, MM algorithm generally converges to a stationary point of the objective function.

The most important component of designing a MM algorithm is to find an appropriate surrogate function $g(x|x^{(t)})$. In general, many surrogate functions may be derived from various inequalities stemming from convexity or concavity, see, e.g., \citet{lange2000optimization} or \citet{hunter2004tutorial22}. One of the most used inequalities to construct a surrogate function is the supporting hyperplane inequality. Suppose $f(x)$ is convex with gradient $\nabla f(x)$, the supporting hyperplane inequality is
\begin{equation}\tag{3.1} \label{eqn3.1}
f(y) \geq f(x)+ \langle \nabla f(x),\; y-x  \rangle.
\end{equation}
Our derivation of the MM  algorithm for the DCOV-based SDR model hinges on the convexity of the two functions mentioned in the next lemma.
\begin{lem}\label{lem1}
	(a) The scalar function $\displaystyle f(x)=x^{\frac 12}-\epsilon \log\left(1+ \frac{  x^{\frac 12}}{\epsilon} \right)  $
	is concave and differentiable in $x>0$ where $\epsilon >0$ is a constant. (b) The matrix function $\displaystyle f({\bf A})=\|{\bf A}c\|_2-\epsilon\log \left( 1 + \frac{\|{\bf A}c\|_2}{\epsilon}  \right)$ is convex and differentiable in the $n\times p$ matrix ${\bf A}$ where $ c\in  \bbR^p$ is a constant vector and $\epsilon >0$ is a constant scalar.
\end{lem}


\subsection{MM Algorithm}

It is often challenging to directly optimize the objective function $(\ref{eqn2.8})$ due to the non-smoothness. One way to tackle the difficulty is to perturb objective function slightly to render it differentiable, then to optimize this differentiable function using a MM algorithm \citep{hunter2005variable,yu2015high}. Motivated by this idea, we introduce a perturbed version $\cV_{n,\epsilon}^2(\bga^{\top}\Z,\Y)$ of the objective function $(\ref{eqn2.8})$ for the DCOV-based SDR model:
\begin{equation}\tag{3.2}\label{eqn3.2}
\displaystyle
\begin{split}
\cV_{n,\epsilon}^2(\bga^{\top}\Z,\Y)
&= \frac{1}{n^2} \sum_{k,l=1}^{n} \left\{  a_{kl}(\bga)  - \epsilon\log \left( 1 + \frac{a_{kl}(\bga)}{\epsilon}  \right)  \right\} B_{kl},                                      \\
&=\frac{1}{n^2} \sum_{k,l=1}^{n} \left\{  \|\bga^{\top}(Z_k-Z_l)  \|_2- \epsilon\log \left( 1 + \frac{\|\bga^{\top}(Z_k-Z_l)\|_2}{\epsilon}  \right)  \right\}  B_{kl}.
\end{split}
\end{equation}
Below we conclude some properties of the perturbed objective function $\cV_{n,\epsilon}^2(\bga^{\top}\Z,\Y)$.
\begin{Pro}\label{Pro1}
	For $\epsilon>0$, (i) $\cV_{n,\epsilon}^2(\bga^{\top}\Z,\Y)$ is a continuous and differentiable DC function and a DC decomposition of it is
	\begin{equation}\tag{3.3}\label{eqn3.3}
	\begin{split}
	\cV_{n,\epsilon}^2(\bga^{\top}\Z,\Y)
	&= \left( \frac{1}{n^2} \sum_{k,l=1}^{n} \left\{  a_{kl}(\bga)  - \epsilon\log \left( 1 + \frac{a_{kl}(\bga)}{\epsilon}  \right)  \right\} B_{kl}I(B_{kl}>0)
	\right) \\
	&- \left( -\frac{1}{n^2} \sum_{k,l=1}^{n} \left\{  a_{kl}(\bga)  - \epsilon\log \left( 1 + \frac{a_{kl}(\bga)}{\epsilon}  \right)  \right\} B_{kl}I(B_{kl}<0)\right) ,
	\end{split}
	\end{equation}
	where $I(\cdot)$ is an indicator function,
	(ii) $\cV_{n,\epsilon}^2(\bga^{\top}\Z,\Y)$ converges to $\cV_{n}^2(\bga^{\top}\Z,\Y)$ uniformly on the Stiefel manifold $\bga\in \St(d,p)$ as $\epsilon$ approaches to zero.
	
\end{Pro}

Now let $\bga^{(t)}$ denote the current estimate, we plan to construct the minorization $g_{\epsilon}(\bga|\bga^{(t)})$ for the perturbed objective function $\cV_{n,\epsilon}^2(\bga^{\top}\Z,\Y)$ based on the DC decomposition $(\ref{eqn3.3})$. The convexity of the function $\displaystyle {\bf A} \mapsto \|{\bf A}c\|_2-\epsilon\log \left( 1 + \frac{\|{\bf A}c\|_2}{\epsilon}  \right)$ implies that
\begin{equation}\notag
\begin{split}
a_{kl}(\bga)  - \epsilon\log \left( 1 + \frac{a_{kl}(\bga)}{\epsilon}  \right) &=\|\bga^{\top}(Z_k-Z_l)\|_2- \epsilon\log \left( 1 + \frac{\|\bga^{\top}(Z_k-Z_l)\|_2}{\epsilon}  \right) \\
&\geq  \|{\bga^{(t)}}^{\top}(Z_k-Z_l) \|_2- \epsilon\log \left( 1 + \frac{\|{\bga^{(t)}}^{\top}(Z_k-Z_l)\|_2}{\epsilon}  \right) \\
& \quad +\big\langle
\frac{(Z_k-Z_l)(Z_k-Z_l)^{\top}\bga^{(t)}}{  \|{\bga^{(t)}}^{\top}(Z_k-Z_l)\|_2+\epsilon },\; \bga-\bga^{(t)}   \big\rangle.
\end{split}
\end{equation}
Multiplying both sides by a nonnegative term $B_{kl}I(B_{kl}>0)$ and averaging over all pairs $(k,l)$ leads to the minorization
\begin{equation}\tag{3.4}\label{eqn3.4}
\begin{split}
& \frac{1}{n^2} \sum_{k,l=1}^{n} \left\{  a_{kl}(\bga)  - \epsilon\log \left( 1 + \frac{a_{kl}(\bga)}{\epsilon}  \right)  \right\} B_{kl}I(B_{kl}>0) \\
&\geq \frac{1}{n^2} \sum_{k,l=1}^{n} \left\{  a_{kl}(\bga^{(t)})  - \epsilon\log \left( 1 + \frac{a_{kl}(\bga^{(t)})}{\epsilon}  \right)  \right\} B_{kl}I(B_{kl}>0) \\
& \quad +  \frac{1}{n^2} \sum_{k,l=1}^{n} \big\langle
\frac{(Z_k-Z_l)(Z_k-Z_l)^{\top}\bga^{(t)}}{  \|{\bga^{(t)}}^{\top}(Z_k-Z_l)\|_2+\epsilon },\; \bga-\bga^{(t)}   \big\rangle B_{kl}I(B_{kl}>0).
\end{split}
\end{equation}
Next focusing on the term $\displaystyle a_{kl}(\bga)  - \epsilon\log \left( 1 + \frac{a_{kl}(\bga)}{\epsilon}  \right)  B_{kl}I(B_{kl}<0)$, we use the fact that $\displaystyle f(x)=x^{\frac 12}-\epsilon \log\left( 1+\frac{ x^{\frac 12}}{\epsilon} \right)$ is concave in $x>0$ to show
\begin{equation}\notag
x^{\frac 12}-\epsilon \log\left( 1+\frac{ x^{\frac 12}}{\epsilon} \right) \leq {x^{(t)}}^{\frac 12}-\epsilon \log\left( 1+\frac{ {x^{(t)}}^{\frac 12}}{\epsilon} \right)+ \frac{x-x^{(t)}}{2\left( {x^{(t)}}^{\frac 12}+\epsilon \right)}.
\end{equation}
Then, we take $x=\|\bga^{\top}(Z_k-Z_l)\|^2_2$ and $x^{(t)}=\|{\bga^{(t)}}^{\top}(Z_k-Z_l)\|^2_2$, the above inequality becomes
\begin{equation}\notag
\begin{split}
& \|\bga^{\top}(Z_k-Z_l)  \|_2- \epsilon\log \left( 1 + \frac{\|\bga^{\top}(Z_k-Z_l)\|_2}{\epsilon}  \right) \\
& \leq \|{\bga^{(t)}}^{\top}(Z_k-Z_l)  \|_2- \epsilon\log \left( 1 + \frac{\|{\bga^{(t)}}^{\top}(Z_k-Z_l)\|_2}{\epsilon}  \right)\\
&\quad +   \frac{  \|\bga^{\top}(Z_k-Z_l)\|^2_2-\|{\bga^{(t)}}^{\top}(Z_k-Z_l)\|^2_2 }{2\left( \|{\bga^{(t)}}^{\top}(Z_k-Z_l) \|_2 + \epsilon \right)}.
\end{split}
\end{equation}
Multiplying both sides by a nonpositive term $B_{kl}I(B_{kl}<0)$ and averaging over all pairs $(k,l)$, we obtain the minorization
\begin{equation}\tag{3.5}\label{eqn3.5}
\begin{split}
& \frac{1}{n^2} \sum_{k,l=1}^{n} \left\{  a_{kl}(\bga)  - \epsilon\log \left( 1 + \frac{a_{kl}(\bga)}{\epsilon}  \right)  \right\} B_{kl}I(B_{kl}<0) \\
&\geq \frac{1}{n^2} \sum_{k,l=1}^{n} \left\{  a_{kl}(\bga^{(t)})  - \epsilon\log \left( 1 + \frac{a_{kl}(\bga^{(t)})}{\epsilon}  \right)  \right\} B_{kl}I(B_{kl}<0) \\
&\quad + \frac{1}{n^2} \sum_{k,l=1}^{n}  \frac{  \|\bga^{\top}(Z_k-Z_l)\|_2^2-\|{\bga^{(t)}}^{\top}(Z_k-Z_l)\|_2^2 }{2\left( \|{\bga^{(t)}}^{\top}(Z_k-Z_l) \|_2 + \epsilon \right)} B_{kl}I(B_{kl}<0).
\end{split}
\end{equation}
Combination of the minorizations $(\ref{eqn3.4})$ and $(\ref{eqn3.5})$ gives the overall minorization
\begin{equation}\tag{3.6}\label{eqn3.6}
\begin{split}
g_{\epsilon}(\bga|\bga^{(t)}) &=  \frac{1}{n^2} \sum_{k,l=1}^{n} \frac{ B_{kl}I(B_{kl}<0) }{2\left( \|{\bga^{(t)}}^{\top}(Z_k-Z_l) \|_2+ \epsilon \right)}\|\bga^{\top}(Z_k-Z_l)\|^2_2 \\
&\quad + \frac{1}{n^2} \sum_{k,l=1}^{n}  \big\langle
\frac{(Z_k-Z_l)(Z_k-Z_l)^{\top}\bga^{(t)}}{  \|{\bga^{(t)}}^{\top}(Z_k-Z_l)\|_2+\epsilon },\; \bga \big\rangle B_{kl}I(B_{kl}>0) + c^{(t)},
\end{split}
\end{equation}
where $c^{(t)}$ is an irrelevant constant. 

To make clear of the surrogate function, we write it in a matrix form. Let $\C$ be a $n \times n$ matrix with every entry $\displaystyle C_{kl}= \frac{ B_{kl}I(B_{kl}<0) }{ \|{\bga^{(t)}}^{\top}(Z_k-Z_l)\|_2 + \epsilon }$ and $\D$ be a $n \times n$ matrix with every entry $\displaystyle D_{kl}= \frac{ B_{kl}I(B_{kl}>0) }{ \|{\bga^{(t)}}^{\top}(Z_k-Z_l)\|_2 + \epsilon }$, then the surrogate function $(\ref{eqn3.6})$ becomes
\begin{equation*}
\begin{split}
g_{\epsilon}(\bga|\bga^{(t)}) &=  \frac{1}{n^2} \sum_{k,l=1}^{n}
\frac{C_{kl}}{2}
\|{\bga}^{\top}(Z_k-Z_l)\|^2_2  \\
&\quad + \frac{1}{n^2} \sum_{k,l=1}^{n}  \big\langle
D_{kl}(Z_k-Z_l)(Z_k-Z_l)^{\top}\bga^{(t)},\; \bga \big\rangle + c^{(t)}.
\end{split}
\end{equation*}
After some algebraic manipulation, we have
\begin{equation*}
\begin{split}
g_{\epsilon}(\bga|\bga^{(t)}) &=  \frac{1}{2} \tr\left( \bga^{\top} \Z \frac{ 2(\diag(\C\1_n)-\C ) }{ n^2 } \Z^{\top} \bga \right)  \\
&\quad + \tr\left( {\bga^{(t)}}^{\top} \Z  \frac{ 2(\diag(\D\1_n)-\D ) }{ n^2 } \Z^{\top} \bga \right) + c^{(t)},
\end{split}
\end{equation*}
where $\1_n$ is a $n\times 1$ column vector having all $n$ elements equal to one and $\diag(a)$ is the $n\times n$ diagonal matrix whose entries are the $n$ elements of the vector $a$.
Let $\displaystyle \Q=\Z  \frac{ 2(\diag(\C\1_n)-\C ) }{ n^2 } \Z^{\top}$ and $\displaystyle \L= \Z \frac{ 2(\diag(\D\1_n)-\D ) }{ n^2 } \Z^{\top}\bga^{(t)}$, the surrogate function $g_{\epsilon}(\bga|\bga^{(t)})$ finally has the form
\begin{equation} \tag{3.7}\label{eqn3.7}
g_{\epsilon}(\bga|\bga^{(t)})=\frac{1}{2} \tr\left( \bga^{\top} \Q \bga \right)+ \tr\left( \bga^{\top} \L \right),
\end{equation}
subject to $\bga\in\St(d,p)$.

Maximizing the surrogate function $g_{\epsilon}(\bga|\bga^{(t)})$ under the constraint drives the loss function uphill. However, due to the existence of the manifold constraint, it is still difficult to accurately solve the subproblem $(\ref{eqn3.7})$ although the objective function is only a quadratic function. In fact, the validity of the ascent property depends only on increasing $g_{\epsilon}(\bga|\bga^{(t)})$ over the Stiefel manifold $\St(d,p)$, not on maximizing $g_{\epsilon}(\bga|\bga^{(t)})$. Similar to \citet{lange1995gradient} and \citet{xu2018majorization}, we propose inexactly maximizing the surrogate function $g_{\epsilon}(\bga|\bga^{(t)})$ by taking a single Newton's step but over the Stiefel manifold $\St(d,p)$. At each iterate $\bga^{(t)}$, we need to solve the following Newton's equation of the problem $(\ref{eqn3.7})$
\begin{equation}\tag{3.8}\label{eqn3.8}
\mbox{Hess}\, g_{\epsilon}(\bga^{(t)})[\bxi]=-\mbox{grad}\,g_{\epsilon}(\bga^{(t)}),
\end{equation}
subject to $\bxi \in {\cal T}_{\bga^{(t)}} \St(d,p)$. After obtaining the Newton's direction $\bxi$ at the current estimate $\bga^{(t)}$, we can update estimate by
\begin{equation*}
\bga^{(t+1)} =\mbox{Retr}_{\bga^{(t)}}(\bxi)={\rm qf}(\bga^{(t)}+\bxi),
\end{equation*}
where ${\rm qf}(\cdot)$ denotes the Q factor of the QR decomposition of the matrix. To safeguard the MM algorithm preserving the ascent property, we can take step-having strategy at every iterate. We call this MM algorithm for soving the DCOV-based SDR model MMRN algorithm and the following Algorithm (\ref{alg1}) summarizes the MMRN algorithm using step-halving based on satisfying the Armijo condition.

\begin{algorithm} \label{alg1}
	\caption{MMRN Algorithm for $(\ref{eqn2.7})$}
	\KwIn{$\X\in\bbR^{p\times n}$, $\Y\in\bbR^{1\times n}$, perturbation constant $\epsilon$}
	Initialize $\bga^{(0)}\in \St(d,p)$, $\alpha\in (0,1)$, $\sigma\in (0,1)$, $t= 0$ \\
	Precompute ${\widehat{\bSigma}_{X}}^{\frac 12}$, $\B=\left( B_{kl}\right)$, $\Z={\widehat{\bSigma}_{X}}^{-\frac 12} \X $ \\
	\Repeat{objective value converges}{
		$\displaystyle  C_{kl} \gets \frac{B_{kl}I(B_{kl}<0) }{ \|{\bga^{(t)}}^{\top}(Z_k-Z_l)\|_2+\epsilon } $,\quad
		$\displaystyle  D_{kl} \gets \frac{B_{kl}I(B_{kl}>0) }{ \|{\bga^{(t)}}^{\top}(Z_k-Z_l)\|_2+\epsilon } $, \quad for any $k,l=1,\ldots,n$    \\[0.1in]
		
		$\displaystyle \Q\gets \Z\frac{2(\diag(\C\1_n)-\C)  }{n^2} \Z^{\top} $, \quad  $\displaystyle \L\gets \Z\frac{2(\diag(\D\1_n)-\D)  }{n^2} \Z^{\top} \bga^{(t)}$\\[0.1in]
		
		Solve the Newton's equation
		\begin{equation}\notag
		\mbox{Hess}\, g_{\epsilon}(\bga^{(t)})[\bxi]=-\mbox{grad}\,g_{\epsilon}(\bga^{(t)}),
		\end{equation}
		for unknown $\bxi \in {\cal T}_{\bga^{(t)}} \St(d,p)$ \\
		$s\gets 1$ \\
		\Repeat{$\displaystyle \cV_{n,\epsilon}^2({{\rm Retr}_{\bga^{(t)}}(s\bxi)}^{\top}\Z,\Y) \geq \cV_{n,\epsilon}^2({\bga^{(t)}}^{\top}\Z,\Y)+\alpha s\|\bxi\|_{\rm F}^2 $}{
			$s\gets \sigma s$			
			
		}
		$\bga^{(t+1)}\gets\mbox{Retr}_{\bga^{(t)}}(s\bxi)$\\
		$t\gets t+1$\\
		
	}
	\KwOut{$\hat{\bga}_{\epsilon}=\bga^{(t+1)},\; \hat{\be}_{\epsilon}={\widehat{\bSigma}_{X}}^{-\frac 12} \hat{\bga}_{\epsilon}$}
\end{algorithm}

\subsection{Solving the Riemannian Newton's equation $(\ref{eqn3.8})$}
The MM algorithm is a well-applicable and simple algorithmic framework for solving DC problems. The key challenge in making the proposed algorithm efficient numerically lies in solving the equation $(\ref{eqn3.8})$. \citet{aihara2017matrix} and \citet{sato2017riemannian} recently proposed an effective way of solving Newton's equation on the Stiefel manifold. The idea of the method is to rewrite original Newton's equation expressed by a system of matrix equations into a standard linear system through the Kronecker product and the vec and veck operators. The resultant linear system can be effectively solved while reducing the dimension of the equation to that of the Stiefel manifold.

Before applying their method to solve the Newton's equation of our subproblem $(\ref{eqn3.8})$ formally, we introduce some useful properties of Kronecker, vec, and veck operators.
\begin{enumerate}
	\item For any $\A\in \bbR^{m\times p}$, $\X \in \bbR^{p\times q}$, and $\B\in\bbR^{q\times n}$, we have
	\begin{equation}\notag
	{\rm vec}(\A\X\B)=\left( \B^{\top}\otimes\A \right){\rm vec}(\X).
	\end{equation}
	\item For any matrix $\U\in {\rm Skew}(d)$, we have
	\begin{equation} \notag
	{\rm vec}(\U)= \D_d {\rm veck}(\U),
	\end{equation}
	and
	\begin{equation} \notag
	{\rm veck}(\U)= \frac 12 \D_d^{\top} {\rm vec}(\U).
	\end{equation}
	Here $\D_d$ is a $d^2 \times d(d-1)/2$ matrix defined by
	\begin{equation*}
	\D_d=\sum_{ d\geq i>j\geq 1  } \left(  \E^{(d^2 \times d(d-1)/2)}_{d(j-1)+i,\,j(d-(j+1)/2)-d+i }  - \E^{(d^2 \times d(d-1)/2)}_{d(i-1)+j,\,j(d-(j+1)/2)-d+i }     \right),
	\end{equation*}
	where $\E^{(p\times q)}_{i,\, j}$ denotes the $p\times q$ matrix that has the $(i,j)$-component equal to $1$ and all other components equal to $0$.
	\item There exists an $n^2\times n^2$ permutation matrix $\T_n$ such that
	\begin{equation}\notag
	{\rm vec}({\bf W}^{\top})=\T_n{\rm vec}({\bf W}), \quad {\bf W} \in \bbR^{n\times n},
	\end{equation}
	where $\T_n=\sum_{i,j=1}^{n}\E_{ij}^{(n\times n)}\otimes \E_{ji}^{(n\times n)} $.
\end{enumerate}
From the above properties, we can easily derive that
\begin{equation} \notag
{\rm vec}({\rm skew}({\bf W}))=\frac 12 (\I_{n^2}-\T_n  ) {\rm vec}({\bf W} ), \quad \mbox{for any } {\bf W} \in \bbR^{n\times n}.
\end{equation}

After these preparations, we begin to solve the Newton's equation $(\ref{eqn3.8})$. For a given $\tilde{\bga}\in \St(d,p)$, the Newton's equation $(\ref{eqn3.8})$ is equivalent to
\begin{equation}\tag{3.9} \label{eqn3.9}
\mbox{Hess}\, g_{\epsilon}(\tilde{\bga})[\bxi]=-\mbox{grad}\,g_{\epsilon}(\tilde{\bga}),
\end{equation}
subject to $\bxi \in {\cal T}_{\tilde{\bga}} \St(d,p)$. Specifically, the gradient of $g_{\epsilon}$ at a point $\tilde{\bga}\in \St(d,p)$ is expressed as
\begin{equation}\tag{3.10}\label{eqn3.10}
\mbox{grad}\,g_{\epsilon}(\tilde{\bga})=\Q\tilde{\bga}+\L-\tilde{\bga}\S,
\end{equation}
and the Hessian acts on $\bxi \in {\cal T}_{\tilde{\bga}} \St(d,p)$ as
\begin{equation}\tag{3.11}\label{eqn3.11}
\mbox{Hess}\, g_{\epsilon}(\tilde{\bga})[\bxi] =  \Q \bxi -\bxi\S-\tilde{\bga}\mbox{sym}\left( {\tilde{\bga}}^{\top} \Q \bxi -{\tilde{\bga}}^{\top}\bxi\S   \right),
\end{equation}
where $\S = \mbox{sym}( {\tilde{\bga}}^{\top} \Q\tilde{\bga}  + {\tilde{\bga}}^{\top} \L )$.
$\bxi \in {\cal T}_{\tilde{\bga}} \St(d,p)$ can be expressed as
\begin{equation}\tag{3.12}\label{eqn3.12}
\bxi = \tilde{\bga} \U  +   \tilde{\bga}_{\perp} \V, \quad \U \in \mbox{Skew}(d), \V \in \bbR^{(p-d)\times d}.
\end{equation}
$\mbox{Hess}\, g_{\epsilon}(\tilde{\bga})[\bxi]   \in {\cal T}_{\tilde{\bga}} \St(d,p)$ can also be written as
\begin{equation}\tag{3.13}\label{eqn3.13}
\mbox{Hess}\, g_{\epsilon}(\tilde{\bga})[\bxi] = \tilde{\bga} \U_{H}  +   \tilde{\bga}_{\perp} \V_{H}, \quad \U_{H} \in \mbox{Skew}(d), \V_{H} \in \bbR^{(p-d)\times d}.
\end{equation}
Substituting the equation $(\ref{eqn3.12})$ into the equation $(\ref{eqn3.11})$ and combining the resultant equation with the equation $(\ref{eqn3.13})$, we can obtain a relationship between $\U_{H}, \V_{H}$ and $\U, \V$. The following proposition gives the relationship.
\begin{Pro}
	Let $\tilde{\bga} \in \St(d,p) $ and $\tilde{\bga}_{\perp}$ be its orthonormal complement. If a tangent vector $\bxi \in {\cal T}_{\tilde{\bga}} \St(d,p) $ is expressed as $(\ref{eqn3.12})$,  then the Hessian $\rm{Hess}\, g_{\epsilon}(\tilde{\bga})[\bxi]$ of the function $(\ref{eqn3.7})$ acts on $\bxi$ as $\rm{Hess}\, g_{\epsilon}(\tilde{\bga})[\bxi] = \tilde{\bga} \U_{H}  +   \tilde{\bga}_{\perp} \V_{H}$ with
	\begin{equation}\tag{3.14}\label{eqn3.14}
	\U_{H}= \rm{skew} \left(  { \tilde{\bga} }^{\top} \Q \tilde{\bga} \U  +  { \tilde{\bga} }^{\top} \Q \tilde{\bga}_{\perp} \V    -\U\S \right),
	\end{equation}
	and
	\begin{equation}\tag{3.15}\label{eqn3.15}
	\V_{H}= {\tilde{\bga}_{\perp}}^{\top} \Q\tilde{\bga} \U + {\tilde{\bga}_{\perp}}^{\top} \Q \tilde{\bga}_{\perp} \V  -\V\S.
	\end{equation}
	\label{pro2}
\end{Pro}

From Equation $(\ref{eqn3.14})$ and $(\ref{eqn3.15})$, we know the Hessian $\rm{Hess} \, g_{\epsilon}(\tilde{\bga})$ at $\tilde{\bga}\in \St(d,p) $ is a linear transformation $\H$ on $\bbR^{K}$ that transforms a $K$-dimensional vector ${ \left( {\rm{veck}(\U)}^{\top}, {{\rm vec}(\V)}^{\top}   \right) }^{\top}$ into ${ \left( {{\rm veck}(\U_{H})}^{\top}, {{\rm vec}(\V_{H})}^{\top}   \right) }^{\top}$. A goal of the method is to obtain the linear transformation $\H$.
\begin{Pro}
	Let $K={\rm dim}(\St(d,p))=d(d-1)/2+(p-d)d$,  there exists a linear transformation $\H$ on $\bbR^{K}$ such that
	\begin{equation*}
	\H \left(
	\begin{array}{c}
	\rm{veck}(\U) \\
	{\rm vec}(\V)
	\end{array}
	\right)=
	\left(
	\begin{array}{c}
	{\rm veck}(\U_{H})\\
	{\rm vec}(\V_{H}) \\
	\end{array}
	\right),
	\end{equation*}
	and the linear transformation $\H$ is given by
	\begin{equation*}
	\H=
	\left(
	\begin{array}{cc}
	\H_{11} &  \H_{12}\\
	\H_{21} &  \H_{22}\\
	\end{array}
	\right),
	\end{equation*}
	where
	\begin{eqnarray*}
		\H_{11} &=& \frac 14  \D^{\top}_d \left[   \I_d \otimes  ({\tilde{\bga}}^{\top}\Q\tilde{\bga}-\S)  + ({\tilde{\bga}}^{\top}\Q\tilde{\bga}-\S)\otimes \I_d  \right] \D_d , \\
		\H_{12} &=& \frac 14 \D_d^{\top}(\I_{d^2}-\T_d )  \left( \I_d\otimes {\tilde{\bga}}^{\top}\Q\tilde{\bga}_{\perp}   \right),\\
		\H_{21} &=& (\I_d\otimes \tilde{\bga}_{\perp}^{\top}\Q\tilde{\bga})\D_d, \\
		\H_{22} &=& \I_d \otimes  {\tilde{\bga}_{\perp}}^{\top}\Q\tilde{\bga}_{\perp} -\S\otimes \I_d.
	\end{eqnarray*}
	\label{pro3}
\end{Pro}

From the Newton's equation $(\ref{eqn3.9})$ together with Equation $(\ref{eqn3.13})$, we have
\begin{equation}\tag{3.16}\label{eqn3.16}
\begin{cases}
\U_{H}&=-\tilde{\bga}^{\top} {\rm grad}\, g_{\epsilon}(\tilde{\bga}),\\
\V_{H}&=-\tilde{\bga}^{\top}_{\perp} {\rm grad}\, g_{\epsilon}(\tilde{\bga}).
\end{cases}
\end{equation}
Applying the veck and vec operators to the equations $(\ref{eqn3.16})$ respectively and using equation $(\ref{eqn3.10})$, we immediately obtain
\begin{equation}\notag
\begin{cases}
{\rm veck}(\U_{H})&=-{\rm veck}  \left( {\rm skew}( {\tilde{\bga}}^{\top}\Q\tilde{\bga}+ {\tilde{\bga}}^{\top}\L ) \right),\\
{\rm vec}(\V_{H})&=-{\rm vec}(\tilde{\bga}^{\top}_{\perp}\Q\tilde{\bga}+\tilde{\bga}^{\top}_{\perp}\L).
\end{cases}
\end{equation}
By Proposition 3, we have a standard linear system
\begin{equation}\notag
\H \left(
\begin{array}{c}
\rm{veck}(\U) \\
{\rm vec}(\V)
\end{array}
\right)=
-
\left(
\begin{array}{c}
{\rm veck}  \left( {\rm skew}( {\tilde{\bga}}^{\top}\Q\tilde{\bga}+ {\tilde{\bga}}^{\top}\L ) \right)\\
{\rm vec}(\tilde{\bga}^{\top}_{\perp}\Q\tilde{\bga}+\tilde{\bga}^{\top}_{\perp}\L) \\
\end{array}
\right).
\end{equation}
If $\H$ is invertible, we can solve the above linear equation as
\begin{equation}\notag
\left(
\begin{array}{c}
\rm{veck}(\U) \\
{\rm vec}(\V)
\end{array}
\right)=
-\H^{-1}
\left(
\begin{array}{c}
{\rm veck}  \left( {\rm skew}( {\tilde{\bga}}^{\top}\Q\tilde{\bga}+ {\tilde{\bga}}^{\top}\L ) \right)\\
{\rm vec}(\tilde{\bga}^{\top}_{\perp}\Q\tilde{\bga}+\tilde{\bga}^{\top}_{\perp}\L) \\
\end{array}
\right).
\end{equation}
In our numerical studies, we have not noticed the case $\H$ is not invertible. After ${\rm veck} (\U)$ and ${\rm vec}(\V)$ are obtained, we can easily reshape $\U \in {\rm Skew}(d)$ and $\V \in \bbR^{(p-d)\times d}$. Therefore, we can calculate the solution of Newton's equation $(\ref{eqn3.9})$ by $\bxi=\tilde{\bga}\U+\tilde{\bga}_{\perp}\V$. Detailed information can be seen in Algorithm $(\ref{alg2})$.

\begin{algorithm} \label{alg2}
	\caption{Solving the Riemannian Newton's equation $(\ref{eqn3.8})$}
	\KwIn{$\Q\in\bbR^{p\times p}$, $\L\in\bbR^{p\times d}$, $\bga^{(t)}\in \bbR^{p\times d}$, $\displaystyle \D_d\in\bbR^{d^2\times \frac{d(d-1)}{2}}$, and $\T_d \in \bbR^{d^2\times d^2}$  }
	Compute ${\bga^{(t)}}_{\perp}$ such that ${\bga^{(t)}}^{\top}\bga^{(t)}_{\perp}=\0$ and ${\bga^{(t)}_{\perp}}^{\top}\bga^{(t)}_{\perp}=\I_{p-d}$ \\
	Compute $\S={\rm sym}( {\bga^{(t)}}^{\top} \Q {\bga^{(t)}}+{\bga^{(t)}}^{\top}\L  )$\\
	Compute the linear transformation ${\H }\in \bbR^{ K \times K}$ by
	\begin{equation*}
	\H=
	\left(
	\begin{array}{cc}
	\H_{11} &  \H_{12}\\
	\H_{21} &  \H_{22}\\
	\end{array}
	\right),
	\end{equation*}
	where
	\begin{eqnarray*}
		\H_{11} &=& \frac 14  \D^{\top}_d \left[   \I_d \otimes  ({\bga^{(t)}}^{\top}\Q\bga^{(t)}-\S)  + ({\bga^{(t)}}^{\top}\Q\bga^{(t)}-\S)\otimes \I_d  \right] \D_d , \\
		\H_{12} &=& \frac 14 \D_d^{\top}(\I_{d^2}-\T_d )  \left( \I_d\otimes {\bga^{(t)}}^{\top}\Q\bga^{(t)}_{\perp}   \right),\\
		\H_{21} &=& (\I_d\otimes {\bga^{(t)}}_{\perp}^{\top}\Q\bga^{(t)})\D_d, \\
		\H_{22} &=& \I_d \otimes  {\bga^{(t)}_{\perp}}^{\top}\Q\bga^{(t)}_{\perp} -\S\otimes \I_d.
	\end{eqnarray*}\\
	Compute ${\rm veck}(\U)$ and ${\rm vec}(\V)$ using
	\begin{equation}\notag
	\left(
	\begin{array}{c}
	\rm{veck}(\U) \\
	{\rm vec}(\V)
	\end{array}
	\right)=
	-\H^{-1}
	\left(
	\begin{array}{c}
	{\rm veck}  \left( {\rm skew}( {\bga^{(t)}}^{\top}\Q\bga^{(t)}+ {\bga^{(t)}}^{\top}\L ) \right)\\
	{\rm vec}({\bga^{(t)}}^{\top}_{\perp}\Q\bga^{(t)}+{\bga^{(t)}}^{\top}_{\perp}\L) \\
	\end{array}
	\right).
	\end{equation}\\
	Construct $\U\in {\rm Skew}(d)$ and $\V \in \bbR^{(p-d)\times d}$ from $\rm{veck}(\U)$ and ${\rm vec}(\V)$  \\
	Compute $\bxi={\bga^{(t)}}\U+\bga^{(t)}_{\perp}\V  $\\
	\KwOut{$\bxi \in \cT_{\bga^{(t)}}\St(d,p) $}
\end{algorithm}

\subsection{Convergence Analysis}
In this section, we construct the convergence property of the proposed algorithm for solving the DCOV-based SDR model. We first show that the sequence $\left\{ \hat{\bga}_{\epsilon}^{(t)} \right\}_{t\geq 0}$ generated by the MMRN algorithm converge to a stationary point of the perturbed function $(\ref{eqn3.2})$.
Then, we show that a maximizer $\hat{\bga}_{\epsilon}$ of the perturbed objective function $(\ref{eqn3.2})$ exhibits a minimal difference from a maximizer $\hat{\bga}$ of the true objective $(\ref{eqn2.8})$ for sufficiently small $\epsilon$.

\begin{Pro}
	Let $\bga \in \St(d,p)$, $\alpha\in(0,1)$, and $\sigma\in (0,1)$, there exists an integer $t>0$ such that
	\begin{equation}\notag
	\cV^2_{n,\epsilon}({{\rm Retr}_{\bga}(\sigma^t\bxi)}^{\top}\Z,\Y) \geq \cV^2_{n,\epsilon}({\bga^{\top}\Z},\Y)+\alpha \sigma^t\|\bxi\|^2_{\rm F},
	\end{equation}
	where $\bxi$ is a solution of $\mbox{Hess}\, g_{\epsilon}(\bga)[\bxi]=-\mbox{grad}\,g_{\epsilon}(\bga)$.
	\label{pro4}
\end{Pro}

We now prove the convergence of our perturbed MM algorithm safeguarded by the Armijo
step-halving strategy.
\begin{Pro}
	For any $\epsilon>0$, the limit point $\hat{\bga}_{\epsilon}$ generated by the Algorithm 1 is a stationary point of $\cV^2_{n,\epsilon}(\bga^{\top}\Z,\Y)$, that is $\mbox{grad}\,\cV^2_{n,\epsilon}({\hat{\bga}_{\epsilon}}^{\top}\Z,\Y) =0.$
	\label{pro5}
\end{Pro}

\begin{Pro} Consider an arbitrary decreasing sequence $\left\{\epsilon_{m}\right\}_{m=1}^{\infty}$ that converges to $0$. Then, any limit point of $\hat{\bga}_{\epsilon_m}$ is a maximizer of $\cV^2_{n}(\bga^{\top}\Z,\Y)$ over the Stiefel manifold, provided that $\left\{\bga|\cV_{n}^2(\bga^{\top}\Z,\Y) = \cV_{n}^2({\hat{\bga}}^{\top}\Z,\Y) \mbox{ and } \bga^{\top}\bga=\I_d\right\}$ is nonempty.
	\label{pro6}
\end{Pro}

Combining Proposition 5 and 6, it is straightforward to see that the MM algorithm generates
solutions that converge to a stationary point of $\cV_{n}^2(\bga^{\top}\Z,\Y)$ as $\epsilon$ decreases to zero.

\begin{thm}
	The sequence of the solutions $\left\{ \hat{\bga}^{(t)}_{\epsilon} \right\}_{t\geq 0}$ generated by the proposed perturbed MM algorithm converges to a maximizer of $\cV_{n}^2(\bga^{\top}\Z,\Y)$ over the Stiefel manifold. Moreover, the sequence of functionals $\left\{ \cV^2_{n,\epsilon}(  \hat{\bga}^{(t)\top}_{\epsilon} \Z,\Y  ) \right\}_{t\geq 0}$ converges to the maximum value of $\cV_{n}^2(\bga^{\top}\Z,\Y)$.
		\label{thm1}
\end{thm}

\section{Extension}
In this section, we will extend the above proposed method to solve sufficient variable selection (SVS) using distance covariance. The DCOV-based SVS method is developed by \citet*{chen2018efficient} through combining DCOV-based SDR with penalty terms, such as LASSO type penalty terms \citep{tibshirani1996regression, yuan2006model,chen2010coordinate} or adaptive LASSO \citep{zou2006adaptive}, to achieve a sparse solution. Specifically, the model is to solve the following problem
\begin{alignat}{1}
\underset{ \be }{\mbox{maximize}} \quad & \cV_{n}^2(\be^{\top}\X,\Y)-\lambda \sum_{i=1}^{p}\theta_i\|\beta_i\|_2 , \tag{4.1} \label{eqn4.1}
\end{alignat}
subject to $\be^{\top} \widehat{\bSigma}_X \be =\I_d $, where $\beta_i$ denotes the $i$-th row vector of $\be$, $\theta_i\geq 0$ serves as the $i$-th penalty weight and $\lambda>0$ is a tunning parameter. Plugging $\bga=\widehat{\bSigma}_{X}^{\frac 12}\be$ and $\Z=\widehat{\bSigma}_{X}^{-\frac 12}\X$ into  the equation $(\ref{eqn4.1})$ together with using equivalent expression $(\ref{eqn2.5})$ for $\cV_{n}^2(\be^{\top}\X,\Y)$, we can transform the objective function $(\ref{eqn4.1})$ to
\begin{equation}\tag{4.2}\label{eqn4.2}
\phi_{\lambda}(\bga)=\frac{1}{n^2}\sum_{k,l=1}^{n}a_{kl}(\bga)B_{kl} -\lambda \sum_{i=1}^{p}\theta_i \rho_i(\bga),
\end{equation}
subject to $\bga\in\St(d,p)$, where $\rho_i(\bga)=\|e_{i}^{\top} \widehat{\bSigma}_{X}^{-\frac 12}\bga
\|_2$ and $e_i$ denotes a column vector with one in the $i$-th position and zero in the others. Correspondingly, a perturbed version $\phi_{\lambda,\epsilon}(\bga)$ of the objective function $(\ref{eqn4.2})$ is given by
\begin{equation}\tag{4.3}\label{eqn4.3}
\displaystyle
\begin{split}
\phi_{\lambda,\epsilon}(\bga)
&= \frac{1}{n^2} \sum_{k,l=1}^{n} \left\{  a_{kl}(\bga)  - \epsilon\log \left( 1 + \frac{a_{kl}(\bga)}{\epsilon}  \right)  \right\} B_{kl}-\lambda\sum_{i=1}^{p} \theta_i\left\{ \rho_i(\bga)-\epsilon\log \left( 1 + \frac{\rho_{i}(\bga)}{\epsilon}  \right)   \right\}, \\
&=\frac{1}{n^2} \sum_{k,l=1}^{n} \left\{  \|\bga^{\top}(Z_k-Z_l)  \|_2- \epsilon\log \left( 1 + \frac{\|\bga^{\top}(Z_k-Z_l)\|_2}{\epsilon}  \right)  \right\}  B_{kl}\\
&\quad -\lambda\sum_{i=1}^{p}\theta_i\left\{  \|e_{i}^{\top} \widehat{\bSigma}_{X}^{-\frac 12}\bga  \|_2 -\epsilon\log\left( 1+\frac{ \|e_{i}^{\top} \widehat{\bSigma}_{X}^{-\frac 12}\bga  \|_2 }{\epsilon} \right)  \right\}.
\end{split}
\end{equation}
Due to the minorization $(\ref{eqn3.7})$ for the first term, it only needs to minorize the penalty function in the equation $(\ref{eqn4.3})$ to obtain a surrogate function of $\phi_{\lambda,\epsilon}(\bga)$. The supporting hyperplane minorization for $\displaystyle -\lambda\theta_i\left\{ x^{\frac 12}-\epsilon\log\left(1+\frac{ x^{\frac 12} }{\epsilon}  \right) \right\}$ is
\begin{equation}\tag{4.4}\label{eqn4.4}
-\lambda\theta_i\left\{ x^{\frac 12}-\epsilon \log\left( 1+\frac{ x^{\frac 12}}{\epsilon} \right) \right\}  \geq   -\lambda\theta_i\left\{ {x^{(t)}}^{\frac 12}-\epsilon \log\left( 1+\frac{ {x^{(t)}}^{\frac 12}}{\epsilon} \right)\right\} + \frac{-\lambda\theta_i(x-x^{(t)})}{2\left( {x^{(t)}}^{\frac 12}+\epsilon \right)}.
\end{equation}
Taking $x=\|e_i^{\top}\widehat{\bSigma}_{X}^{-\frac 12} \bga \|^2_2$ and $x^{(t)}= \|e_i^{\top}\widehat{\bSigma}_{X}^{-\frac 12} \bga^{(t)} \|^2_2 $, and summing over $i=1,\ldots,p$ give the minorization for penalty function $-\lambda \sum_{i=1}^{p}\theta_i \rho_i(\bga)$, i.e., 
\begin{equation}\tag{4.5}\label{eqn4.5}
-\lambda \sum_{i=1}^{p}\theta_i \rho_i(\bga) \geq \sum_{i=1}^{p} \frac{-\lambda\theta_i\|e_i^{\top}\widehat{\bSigma}_{X}^{-\frac 12} \bga \|^2_2}{2(\|e_i^{\top}\widehat{\bSigma}_{X}^{-\frac 12} \bga^{(t)} \|_2+\epsilon)}+c,
\end{equation}
where $c$ is an irrelevant constant. After some algebraic manipulation, we have
\begin{equation}\tag{4.6}\label{eqn4.6}
\sum_{i=1}^{p} \frac{-\lambda\theta_i\|e_i^{\top}\widehat{\bSigma}_{X}^{-\frac 12} \bga \|^2_2}{2(\|e_i^{\top}\widehat{\bSigma}_{X}^{-\frac 12} \bga^{(t)} \|_2+\epsilon)}
=\frac 12 \tr\left(  \bga^{\top} \widehat{\bSigma}_{X}^{-\frac 12} \diag({ \Lambda}) \widehat{\bSigma}_{X}^{-\frac 12}  \bga \right),
\end{equation}
where $\displaystyle  \Lambda=\left( \frac{-\lambda\theta_1 }{\|e_1^{\top}\widehat{\bSigma}_{X}^{-\frac 12} \bga^{(t)} \|_2+\epsilon},\ldots,   \frac{-\lambda\theta_p }{\|e_p^{\top}\widehat{\bSigma}_{X}^{-\frac 12} \bga^{(t)} \|_2+\epsilon} \right)^{\top}$ is a $p\times 1$ column vector. Combining the minorizations $(\ref{eqn3.7})$ and $(\ref{eqn4.6})$ gives the overall minorization
\begin{equation}\tag{4.7}\label{eqn4.7}
g_{\lambda,\epsilon}(\bga|\bga^{(t)})=\frac 12 \tr\left(  \bga^{\top} \left[ \Q+\widehat{\bSigma}_{X}^{-\frac 12} \diag({\Lambda}) \widehat{\bSigma}_{X}^{-\frac 12} \right]   \bga \right)+\tr(\bga^{\top}\L).
\end{equation}
Note that the form of surrogate function $(\ref{eqn4.7})$ for the DCOV-based SVS model is the same as the surrogate function $(\ref{eqn3.7})$ for the DCOV-based SDR model. Thus, we can use the same method for solving the DCOV-based SVS model.

\section{Numerical Studies}
We compare our proposed unified algorithm for solving both DCOV-based SDR and DCOV-based SVS to their corresponding existing algorithms, focusing on computational cost. Since the method in \citet*{chen2018efficient} solving DCOV-based SVS combines SQP and local quadratic approximation \citep[LQA;][]{fan2001variable}, we denote it to SQP+LQA for convenience. SQP and SQP+LQA  in all of the simulation studies use the default setups in the original work to guarantee accuracy. In the MMRN, we set the stepsize multiplicative factor $\sigma=0.5$ and perturbation constant $\epsilon=10^{-10}$ to avoid machine precision error. Besides, we set $\alpha=10^{-20}$ to lead fewer number of line search steps. The MMRN algorithm terminates at the $t$-th step when the relative error of the objective function at the $t$-th step computed by $|f(\bga^{(t)})-f(\bga^{(t-1)})|/|f(\bga^{(t-1)})|$ becomes smaller than $10^{-7}$ or the iteration number $t$ exceeds $1000$. Here the function $f$ denotes the objective functions in DCOV-based SDR and SVS. All algorithms use the solutions from existing dimension reduction methods such as SIR or DR as the initial value. All codes are implemented in \verb|Matlab| and run on a standard PC (Intel Core i9-8950HK CPU (2.90 GHz) and 32 GB RAM). For specific details about the implementation of our proposal, please refer to \url{https://github.com/runxiong-wu/MMRN}.

\subsection{Simulation for DCOV-based SDR}
We use the same simulation settings as in \citet*{sheng2016sufficient} to illustrate the performance comparison of the MMRN algorithm and the SQP algorithm in solving DCOV-based SDR models. There are three different models and two sample size configurations $(n,p)=(100,6)$ and $(500,20)$. Let $\epsilon$, $\epsilon_1$, and $\epsilon_2$ be independent standard normal random variables, the three models are:
\begin{equation*}
\begin{array}{l}
\text { (A) } \quad Y= (\beta_1^{\top} X )^{2}+(\beta_2^{\top} X )+0.1\epsilon, \\
\text { (B) } \quad Y=\operatorname{sign}\left(2 \beta_1^{\top} X +\epsilon_{1}\right) \times \log \left|2 \beta_2^{\top} X +4+\epsilon_{2}\right|,  \\
\text { (C) } \quad Y=\exp ( \beta_3^{\top} X ) \epsilon,
\end{array}
\end{equation*}
where $\beta_1,\beta_2,$ and $\beta_3$ are $p$-dimensional vectors with their first six components being $(1,0,0,0,0,0)^{\top},(0,1,0,0,0,0)^{\top},$ and $(1,0.5,1,0,0,0)^{\top}$ and the last $p-6$ components being $0$ if $p>6$. Each model has three different kinds of $ X=(x_1,\ldots,x_p)^{\top}$: Part (1), standard normal predictors $X\sim N(0,\I_p)$; Part (2), nonnormal predictors; and Part (3), discrete predictors. Specific predictors setups for Part (2) and Part (3) in each model are summarized in Table 1.

\begin{table}[width=.9\linewidth,cols=3,pos=h]
	\caption{Setups for Part (2) and Part (3). Here iid means independent identically distributed.}
	\label{tabel1}
	\begin{tabular*}{\tblwidth}{@{} LLL@{} }
		\toprule
		      & Part (2) & Part (3) \\
		\midrule
		Model A &  $\displaystyle \left\{ \frac{x_i+2}{5}\right\}_{i=1}^{p} \stackrel{\rm iid}{\sim}\mbox{Beta}(0.75, 1)$  & $ \left\{ x_i\right\}_{i=1}^{p} \stackrel{\rm iid}{\sim}\mbox{Poisson}(1)$ \\
	    Model B & $\left\{ x_i\right\}_{i=1}^{p} \stackrel{\rm iid}{\sim}\mbox{Uniform}(-2, 2)$ & $\left\{ x_i\right\}_{i=1}^{p} \stackrel{\rm iid}{\sim}\mbox{Binomial}(10, 0.1)$ \\
    	Model C & $\displaystyle \left\{ \frac{x_i+1}{2}\right\}_{i=1}^{p} \stackrel{\rm iid}{\sim} \mbox{Beta}(1.5, 1)$ &
		$ \left\{ x_i \right\}_{i \not = 6} \stackrel{\rm iid}{\sim} \mbox{Poisson}(1) \mbox{ and }
		x_6 \sim \mbox{Binomial}(10, 0.3) $ \\
		\bottomrule
	\end{tabular*}
\end{table}

Each simulation scenario repeats 100 times. At each time, we use the following distance to measure the accuracy of the estimator $\hat{\be}$
\begin{equation}\notag
\Delta_m(P_{\hat{\be}}, P_{\be})=\|P_{\hat{\be}}-P_{\be} \|,
\end{equation}
where $\be$ is a basis of the true central subspace, $P_{\hat{\be}}$ and $P_{\be}$ are the respective projections of $\hat{\be}$ and $\be$, and $\|\cdot\|$ is the maximum singular value of a matrix. The smaller the $\Delta_m$ is, the more accuracy the estimator is. We report the mean and the standard error of $\Delta_m$'s and CPU times in Table \ref{tabel2}. We can observe that both the SQP algorithm and the MMRN algorithm have satisfactory performance in terms of estimation accuracy, but the MMRN algorithm takes less time than the SQP algorithm. For part (3) of model A at $n=500$ and $p=20$, the MMRN algorithm takes about 2 seconds on average while the SQP algorithm averages more than 50 seconds. It is approximately 25 times faster. Also, the MMRN algorithm is more stable than the SQP algorithm since the standard deviation of the running time is less. Overall, the MMRN algorithm has almost the same performance as the SQP algorithm across various models, but with less time.

\begin{table}[width=\linewidth,cols=7,pos=!htbp]	
	\caption{Simulation results under the same settings as in \citet*{sheng2016sufficient}. The mean (standard error), averaged over $100$ datasets, are reported.}
	\label{tabel2}
	\begin{tabular*}{\tblwidth}{@{}LLLLLLL@{} }
	\toprule
	\multirow{2}{*}{$(n,\;p)$}&\multirow{2}{*}{Model}&\multirow{2}{*}{Part}&\multicolumn{2}{L}{  SQP }&\multicolumn{2}{L}{ MMRN }\cr			
	\cmidrule(rr){4-5} \cmidrule(rr){6-7}
    &&&  $\bar{\Delta}_m$ &Time (sec)& $\bar{\Delta}_m$ &Time (sec)\cr
	\midrule
  $n=100,\; p=6$&  A   &  (1) &0.19(0.06)&0.52(0.16)&0.19(0.06)&0.08(0.03)\\
                &	  &  (2)  &0.19(0.06)&0.55(0.09)&0.19(0.06)&0.07(0.02)\\
             	&      &  (3) &0.00(0.01)&1.18(0.26)&0.00(0.01)&0.12(0.08)\\
	            &  B   &  (1) &0.29(0.10)&0.49(0.20)&0.29(0.10)&0.18(0.09)\\
             	&      &  (2) &0.22(0.07)&0.44(0.08)&0.22(0.07)&0.10(0.03)\\
             	&      &  (3) &0.28(0.18)&0.48(0.17)&0.27(0.18)&0.13(0.10)\\
             	&  C   &  (1) &0.20(0.07)&0.38(0.19)&0.20(0.07)&0.16(0.06)\\
             	&      &  (2) &0.31(0.12)&0.33(0.08)&0.30(0.10)&0.25(0.13)\\
             	&      &  (3) &0.22(0.10)&0.39(0.11)&0.22(0.10)&0.11(0.05)\\
 	\midrule			
 $n=500,\; p=20$&  A   &  (1) &0.16(0.02)&11.41(1.84)&0.16(0.02)&1.27(0.12)\\
	            &	  &  (2)  &0.17(0.03)&13.47(1.96)&0.17(0.03)&1.31(0.14)\\
				&      &  (3) &0.00(0.00)&53.61(4.84)&0.00(0.00)&2.02(0.58)\\
				&  B   &  (1) &0.24(0.04)&10.26(1.63)&0.24(0.04)&3.03(0.37)\\
				&      &  (2) &0.19(0.03)&10.56(2.40)&0.19(0.03)&1.92(0.20)\\
				&      &  (3) &0.18(0.07)&14.72(3.64)&0.18(0.07)&2.24(0.47)\\
				&  C   &  (1) &0.15(0.03)&9.64(0.96)&0.15(0.03)&4.13(0.67)\\
				&      &  (2) &0.24(0.04)&11.20(1.16)&0.24(0.04)&10.59(3.16)\\
				&      &  (3) &0.14(0.03)&12.29(1.37)&0.14(0.03)&3.34(0.55)\\		
				\bottomrule
    \end{tabular*}	
\end{table}

\begin{figure}[pos=!htbp]
	\centering
	\subfigure[Model A part (1)]{
		\begin{minipage}[t]{0.5\linewidth}
			\centering
			\includegraphics[width=\textwidth]{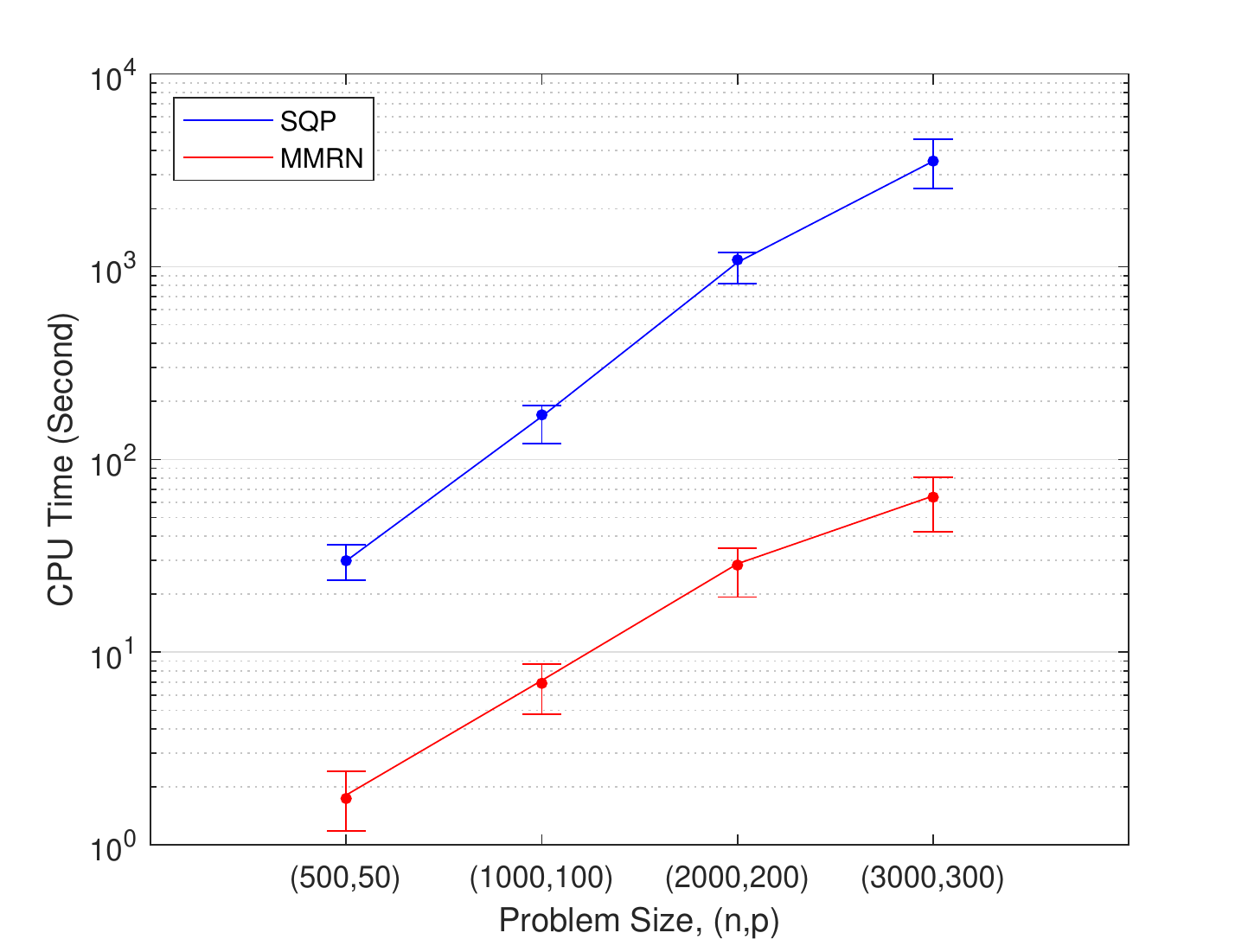}
		\end{minipage}%
	}%
	\subfigure[Model B part (1)]{
		\begin{minipage}[t]{0.5\linewidth}
			\centering
			\includegraphics[width=\textwidth]{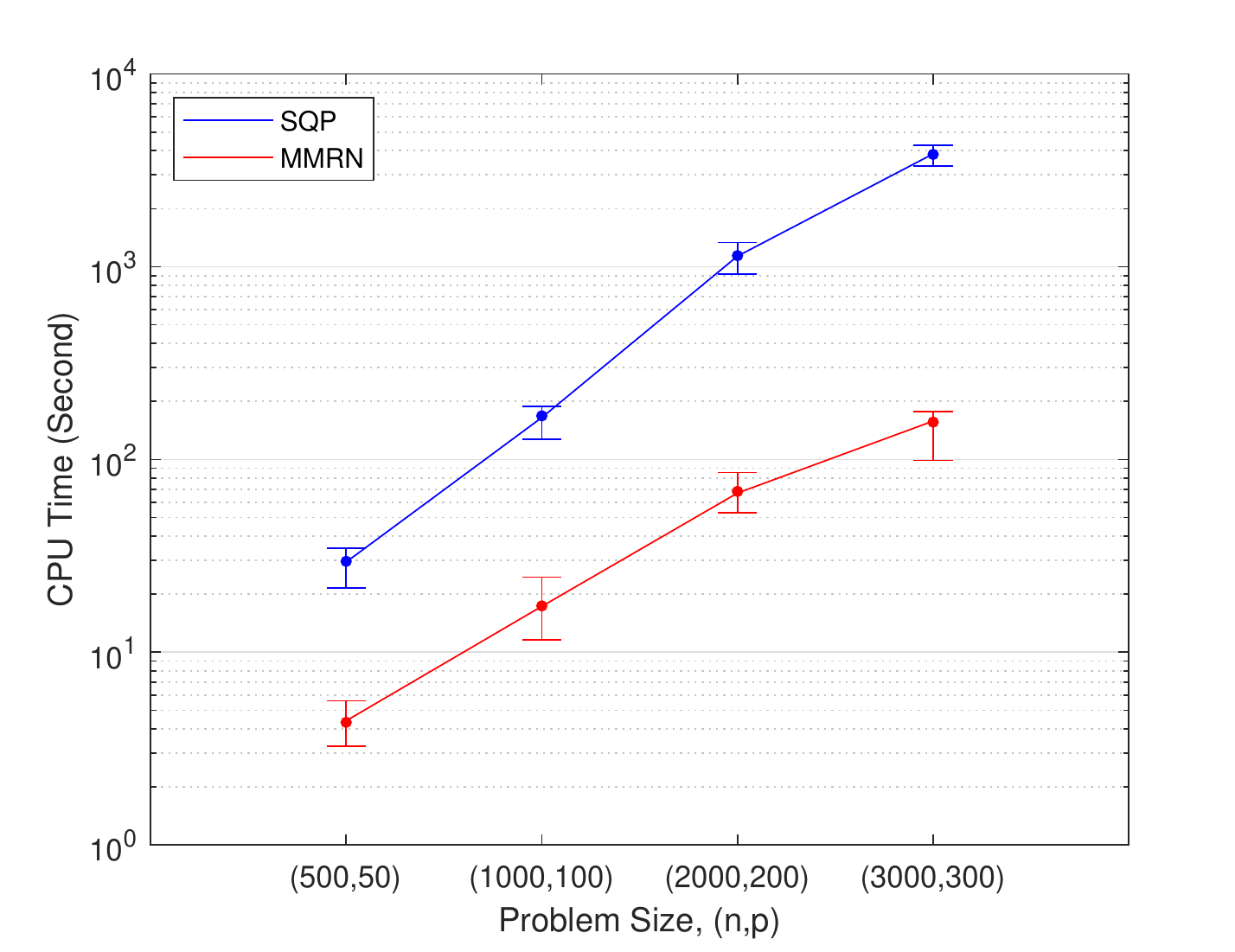}
		\end{minipage}%
	}%
	\\
	\subfigure[Model C part (1)]{
		\begin{minipage}[t]{0.5\linewidth}
			\centering
			\includegraphics[width=\textwidth]{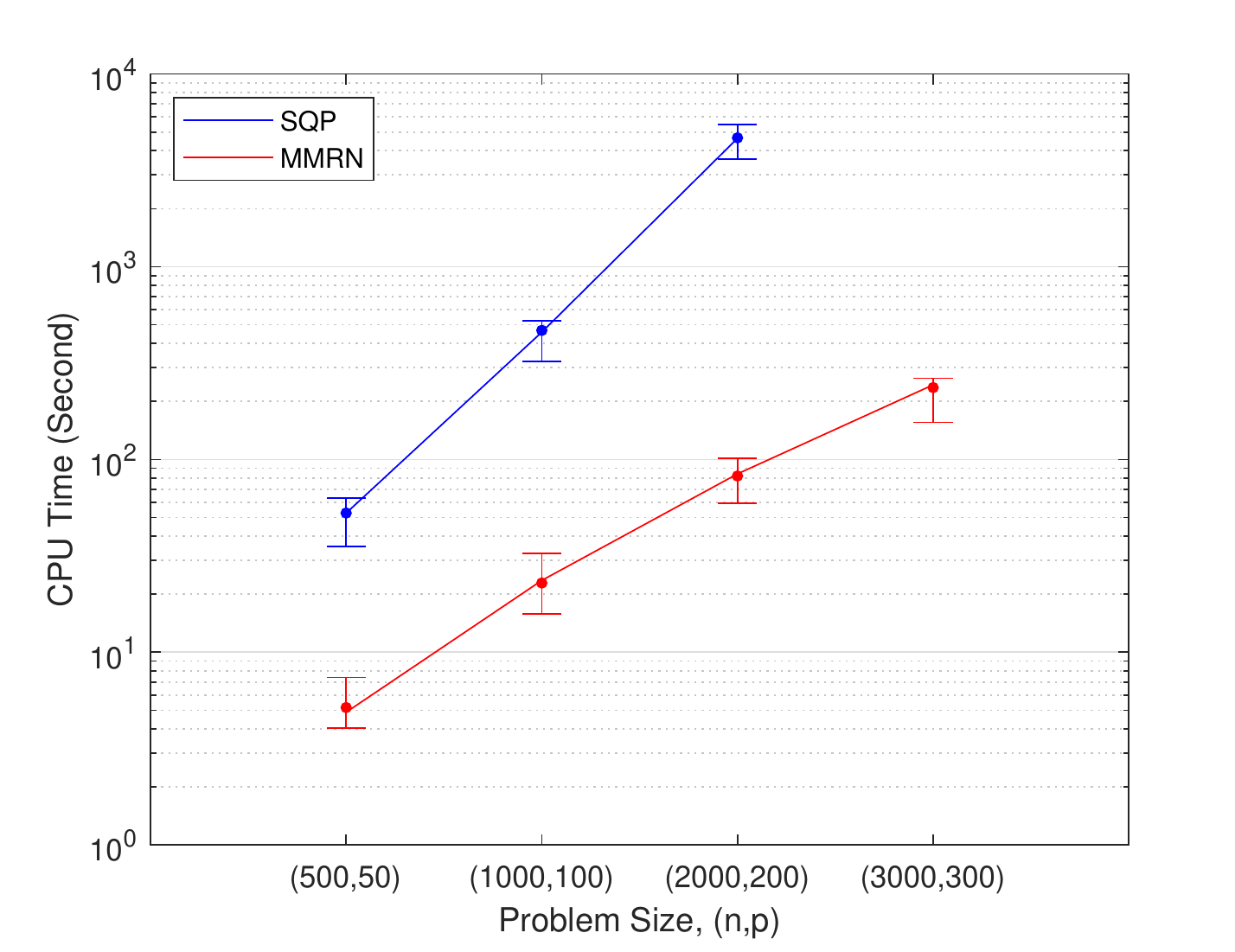}
		\end{minipage}
	}
	\centering
	\caption{ Computational performance comparison on large problem size for three different models with standard normal predictors. The mean of the CPU time averaged over 20 datasets are reported. There was no significant difference of the two methods in the estimation accuracy. Therefore, estimation accuracy is not displayed the graph.}
	\label{fig2}
\end{figure}

To test the performance of our proposed MMRN algorithm in large datasets, we use four different levels for sample size configuration, $(n,p)$: $(500, 50)$, $(1000, 100)$, $(2000, 200)$, and $(3000, 300)$. Here, we only consider the cases with the standard predictors and generate 20 datasets for each study. Figure \ref{fig2} displays a graph of the average runtime for each algorithm under the different problem sizes considered. We can see that our proposed algorithm can outperform the SQP algorithm even in large datasets. Note that we did not run the SQP algorithm on sample size $(n,p)=(3000,300)$ for model C with standard predictors since it will take much time ($> 7$ hours once) to solve the problem.

\subsection{Simulation for DCOV-based SVS}
This part compares the performance of our proposed MMRN algorithm and the SQP+LQA algorithm in solving DCOV-based SVS models. We consider two sample size configurations $(n,p)=(60,24)$ and $(120,24)$ and generate 100 datasets for each simulation. To assess how well the algorithms select variables, we define the true positive rate TPR as the proportion of correctly identified active predictors, and the false positive rate FPR as the proportion of irrelevant predictors that are incorrectly identified to be active. When computing the TPR and FPR in practice, the estimate obtained by the MMRN algorithm is truncated by zeroing out its entries whose magnitude is smaller than $10^{-7}$. In addition, we use the Bayesian information criterion (BIC) to select the tuning parameters, see., \citet*{chen2018efficient}.

We conduct the following simulation studies with the same model settings as the scenarios $n > p$ in \citet*{chen2018efficient}.
\begin{itemize}[leftmargin= 60 pt]
	\item[{\it Study 1.}] A nonlinear regression model with four active predictors:
	\begin{equation}\notag
	Y=(\beta_1^{\top}X+0.5)^2+0.5\epsilon,
	\end{equation}
	where $\epsilon\sim N(0,1)$ and $X\sim N(0, \bSigma)$ with $\Sigma_{ij}=0.5^{|i-j|}$ for $1\leq i,j\leq 24$. The central subspace is spanned by the vectors $\beta_1=(0.5,0.5,0.5,0.5,0_{p-4})^{\top}$.
	\item[{\it Study 2.}] A nonlinear regression model with two active predictors:
	\begin{equation}\notag
	Y=\frac{\beta_1^{\top}X}{ 0.5+(\beta_2^{\top}X+1.5)^2 }+0.2\epsilon,
	\end{equation}
	where $\epsilon\sim N(0,1)$ and $X\sim N(0, \bSigma)$ with $\Sigma_{ij}=0.5^{|i-j|}$ for $1\leq i,j\leq 24$. The central subspace is spanned by the vectors $\beta_1=(1,0,0_{p-2})^{\top}$ and $\beta_2=(0,1,0_{p-2})^{\top}$.
	\item[{\it Study 3.}] A nonlinear regression model with four active predictors:
	\begin{equation}\notag
	Y=(\beta_1^{\top}X)^2+|\beta_2^{\top}X|+0.5\epsilon,
	\end{equation}
	where $\epsilon\sim N(0,1)$. The predictor $X=(x_1,\ldots,x_{24})^{\top}$ is defined as follows: the last $23$ components $(x_2,\ldots,x_{24})^{\top} \sim N(0, \bSigma)$ with $\Sigma_{ij}=0.5^{|i-j|}$ for $1\leq i,j\leq 23$ and the first component $x_1=|x_2+x_3|+\xi$, where $\xi\sim N(0,1)$. The central subspace is spanned by the vectors $\beta_1=(0.5,0.5,0.5,0.5,0_{p-4})^{\top}$ and $\beta_2=(0.5,-0.5,0.5,-0.5,0_{p-4})^{\top}$.
	\item[{\it Study 4.}] A multivariate response model with four active predictors:
	\begin{equation}\notag
	\left\{
	\begin{aligned}
	Y_1 &=\beta_1^{\top} X+\epsilon_{1}, \\
	Y_2 &=(\beta_2^{\top} X+0.5)^2+\epsilon_{2}, \\
	\end{aligned}
	\right.
	\end{equation}
	where $\epsilon_{1}, \epsilon_{2} \stackrel{\rm iid}{\sim} N(0,1)\mbox{ and } X \sim N(0, \Sigma)$ with $\Sigma_{ij}=0.5^{|i-j|}$ for $1\leq i,j\leq 24$. The central subspace is spanned by the vectors $\beta_1=(0.5,0.5,0.5,0.5,0_{p-4})^{\top}$ and $\beta_2=(0.5,-0.5,0.5,-0.5,0_{p-4})^{\top}$.
\end{itemize}
Table \ref{tabel3} gives the simulation results. The MMRN algorithm is much less time-consuming than the SQP+LQA algorithm to achieve the same or even slightly better effect in terms of TPR and FPR. Especially in Study 2 and Study 4, we can observe that the performance of MMRN algorithm in TPR and FPR is better than SQP + LQA, but its speed is nearly 100 times faster.

\begin{table}[width=.9\linewidth,cols=8,pos=!htpb]	
	\centering
	\caption{Simulation results under the same settings as in \citet*{chen2018efficient}. The mean, averaged over $100$ datasets, are reported.}
	\label{tabel3}
	\begin{tabular*}{\tblwidth}{@{}  LLLLLLLL@{}}
	\toprule
	\multirow{2}{*}{}&\multirow{2}{*}{}&\multicolumn{3}{L}{  SQP+LQA }&\multicolumn{3}{L}{ MMRN }\cr			
	\cmidrule(rr){3-5} \cmidrule(rr){6-8}
	& & TPR &FPR &Time (sec)& TPR &FPR &Time (sec)\cr
    \midrule
Study 1&$n=60$ &0.695&0.063&385.5&0.685&0.077&12.1 \cr
       &$n=120$&0.990&0.004&532.9&0.988&0.002&27.6\cr	
Study 2&$n=60$ &0.770&0.031&1051.2&0.870&0.016&5.4 \cr
       &$n=120$&0.930&0.010&1518.3&0.975&0.004&9.4\cr	
Study 3&$n=60$ &0.715&0.010&1122.4&0.725&0.002&11.9 \cr
       &$n=120$&0.785&0.002&1746.3&0.785&0.001&26.8\cr
Study 4&$n=60$ &0.655&0.029&1293.8&0.700&0.011&12.9\cr
	   &$n=120$&0.905&0.009&1778.4&0.930&0.007&30.0\cr	        							
	\bottomrule
	\end{tabular*}	
\end{table}

\subsection{Real Data Analysis}
In this part, we revisit the Boston housing data from \citet{HARRISON197881}; \citet{zhou2008}; and \citet{chen2018efficient} to compare our proposed MMRN algorithm with the SQP+LQA algorithm. Following the previous studies, we remove those observations with crime rate greater than $3.2$. The trimmed Boston housing data contains $374$ observations with the response variable $Y$ being the median value of owner-occupied homes in each of the $374$ census tracts in the Boston Standard Metropolitan Statistical Areas. There are $13$ predictors, which correspond to per capita crime rate by town; proportion of residential land zoned for lots over $25,000$ sq.ft; proportion
of nonretail business acres per town; Charles River dummy variable; nitric oxides concentration; average number of rooms per dwelling; proportion of owner-occupied units built prior to $1940$; weighted distances to five Boston employment centers; index of accessibility to radial high-
ways; full-value property-tax rate; pupil-teacher ratio by town; proportion of blacks by town; percentage of lower status of the population. It has been found two directions are good to estimate the central subspace. After these preparations, we fit the DCOV-based SVS model using the SQP+LQA algorithm and the MMRN algorithm. There is little difference on their predictive performance, but the computing time for these two methods is very different. As we observe, the total optimization time is about $2464$ seconds for the SQP+LQA algorithm and about $52.34$ seconds for the MMRN algorithm. Our algorithm is approximately $47$ times faster than the competitor.

\section{Conclusion}
In the article, we notice that the empirical distance covariance can have a difference of convex functions decomposition. Based on this observation, we leverage the MM principle to design powerful and versatile algorithms uniformly for DCOV-based SDR and DCOV-based SVS models. The proposed algorithms take one single Riemannian Newton's step at each iterate to tackle the Manifold constraints.  The simulation studies show our proposed algorithms are highly efficient and very stable even in large $n$ and large $p$ scenarios. Furthermore, we establish the convergence property of our proposed algorithms under mild conditions.

As a possible future work, we plan to design a new algorithm with the aim to handle the large $p$ small $n$ scenarios directly rather than incorporate it in the framework of sequential SDR \citep{yin2015sequential}.

\section*{Acknowledgments}
We gratefully thank the Editor, the Associate Editor and two referees for all the questions, constructive comments and suggestion. This work is supported by SUSTech startup funding. 

\appendix
\section*{Appendix}

\noindent {\bf Proof of Lemma \ref{lem1}.} For part (a), for any $x>0$, we have the second derivative $\displaystyle f''(x)= - \frac{1}{4(\epsilon+\sqrt{x})^2  \sqrt{x} } \leq 0$\ and immediately obtain that $f(x)$ is concave in $x>0$.

For part (b), recall that $\displaystyle g(x)=x-\epsilon\log\left( 1+\frac{x}{\epsilon}  \right)  $ is convex and increasing in $x>0$, and $h({\bf A})=\|{\bf A}c\|_2$ is convex in ${\bf A} \in \bbR^{n\times p}$. By the composition property we know function $g(h({\bf A}))$ is convex. Thus, we complete our proof.

\noindent {\bf Proof of Proposition \ref{Pro1}.} For part (i), it only needs to prove that $\displaystyle a_{kl}(\bga)-\epsilon \log\left(  1+\frac{a_{kl}(\bga)}{\epsilon} \right)$ is convex with respect to $\bga$. This proof follows immediately from the part (b) of Lemma \ref{lem1} when you take $\A=\bga^{\top}$ and $c= Z_k-Z_l$.

For part (ii), recall that
\begin{equation*}
\begin{aligned}
0\leq \cV_{n}^2(\bga^{\top}\Z,\Y)-\cV_{n,\epsilon}^2(\bga^{\top}\Z,\Y)&=\frac{1}{n^2}\sum_{k,l=1}^{n}\epsilon\log\left( 1+\frac{\|\bga^{\top}(Z_k-Z_l)\|_2 }{\epsilon} \right)\\
&\leq \frac{1}{n^2}\sum_{k,l=1}^{n}\epsilon\log\left( 1+\frac{ \sup_{\bga\in \St(d,p)} \|\bga^{\top}(Z_k-Z_l)\|_2 }{\epsilon} \right)
\end{aligned}
\end{equation*}
The suprema in the rightmost side are achieved and finite because $\St(d,p)$ is bounded. Then, the rightmost side monotonically decreases to 0 as $\epsilon$ goes to 0.

\noindent {\bf Proof of Proposition \ref{pro2}.}
Multiplying the equation $(\ref{eqn3.11})$ by ${\tilde{\bga}}^{\top}$ from the left and using the relations ${\tilde{\bga}}^{\top}\tilde{\bga}=\I_d$ and ${\tilde{\bga}}^{\top}\tilde{\bga}_{\perp}=\0$ yield
\begin{equation}\tag{A.1}\label{eqnA.1}
\begin{aligned}
\U_{H} &= {\tilde{\bga}}^{\top} \mbox{Hess}\, g_{\epsilon}(\tilde{\bga})[\bxi], \\
&={\tilde{\bga}}^{\top} \Q\bxi-{\tilde{\bga}}^{\top}\bxi\S-\mbox{sym}({\tilde{\bga}}^{\top}\Q\bxi-{\tilde{\bga}}^{\top}\bxi\S), \\
&=\mbox{skew}({\tilde{\bga}}^{\top} \Q\bxi-{\tilde{\bga}}^{\top}\bxi\S).
\end{aligned}
\end{equation}
Similarly, we multiply the equation $(\ref{eqn3.11})$ by ${\tilde{\bga}_{\perp}}^{\top}$ from the left to obtain
\begin{equation}\tag{A.2}\label{eqnA.2}
\begin{aligned}
\V_{H} &= {\tilde{\bga}_{\perp}}^{\top} \mbox{Hess}\, g_{\epsilon}(\tilde{\bga})[\bxi], \\
&={\tilde{\bga}_{\perp}}^{\top} \Q\bxi-{\tilde{\bga}_{\perp}}^{\top}\bxi\S.
\end{aligned}
\end{equation}
Substituting the expression $(\ref{eqn3.12})$ of $\bga$ into $(\rm\ref{eqnA.1})$ and $(\rm\ref{eqnA.2})$, we can immediately obtain equations $(\ref{eqn3.14})$ and $(\ref{eqn3.15})$.

\noindent {\bf Proof of Proposition \ref{pro3}.}
From equations $(\ref{eqn3.14})$ and $(\ref{eqn3.15})$ together with the properties of these operators and $\U^{\top}=-\U$, ${\rm veck}(\U_{H})$ and ${\rm vec}(\V_{H})$ are calculated as follows:
\begin{equation}\notag
\begin{aligned}
{\rm veck}(\U_{H}) &=  \frac 12 \D_d^{\top} {\rm vec} (\U_{H}) \\
&= \frac 12 \D_d^{\top} {\rm vec} \left( {\rm skew} ( {\tilde{\bga}}^{\top}\Q\tilde{\bga}\U+{\tilde{\bga}}^{\top}\Q\tilde{\bga}_{\perp}\V-\U\S  ) \right)  \\
&=\frac 14 \D_d^{\top}(\I_{d^2}-\T_d ) {\rm vec} \left(  {\tilde{\bga}}^{\top}\Q\tilde{\bga}\U+{\tilde{\bga}}^{\top}\Q\tilde{\bga}_{\perp}\V-\U\S  \right)\\
&= \frac 14 \D_d^{\top} {\rm vec} \left(  {\tilde{\bga}}^{\top}\Q\tilde{\bga}\U-\U^{\top}{\tilde{\bga}}^{\top}\Q\tilde{\bga}-\U\S+\S\U^{\top}   \right) \\
&\quad\;  +  \frac 14 \D_d^{\top}(\I_{d^2}-\T_d ) {\rm vec} \left( {\tilde{\bga}}^{\top}\Q\tilde{\bga}_{\perp}\V  \right)\\
&=\frac 14 \D_d^{\top} {\rm vec} \left(  {\tilde{\bga}}^{\top}\Q\tilde{\bga}\U+\U{\tilde{\bga}}^{\top}\Q\tilde{\bga}-\U\S-\S\U   \right) \\
&\quad\;  +  \frac 14 \D_d^{\top}(\I_{d^2}-\T_d ) {\rm vec} \left( {\tilde{\bga}}^{\top}\Q\tilde{\bga}_{\perp}\V  \right)\\
&=\frac 14 \D_d^{\top}  \left[  \I_d \otimes  ({\tilde{\bga}}^{\top}\Q\tilde{\bga}-\S)  + ({\tilde{\bga}}^{\top}\Q\tilde{\bga}-\S)\otimes \I_d   \right]   \D_d {\rm veck}(\U)\\
&\quad\; + \frac 14 \D_d^{\top}(\I_{d^2}-\T_d )  \left( \I_d\otimes {\tilde{\bga}}^{\top}\Q\tilde{\bga}_{\perp}   \right) {\rm vec} (\V)\\
&=\H_{11}{\rm veck}(\U)+\H_{12}{\rm vec} (\V),
\end{aligned}
\end{equation}
and
\begin{equation}\notag
\begin{aligned}
{\rm vec}(\V_{H}) &={\rm vec}  \left(  \tilde{\bga}_{\perp}^{\top}\Q\tilde{\bga}\U+\tilde{\bga}_{\perp}^{\top}\Q\tilde{\bga}_{\perp}\V-\V\S    \right)   \\
&=(\I_d\otimes \tilde{\bga}_{\perp}^{\top}\Q\tilde{\bga}){\rm vec}(\U)+(\I_d \otimes  {\tilde{\bga}_{\perp}}^{\top}\Q\tilde{\bga}_{\perp} -\S\otimes \I_d){\rm vec}(\V)\\
&=(\I_d\otimes \tilde{\bga}_{\perp}^{\top}\Q\tilde{\bga})\D_d{\rm veck}(\U)+(\I_d \otimes  {\tilde{\bga}_{\perp}}^{\top}\Q\tilde{\bga}_{\perp} -\S\otimes \I_d){\rm vec}(\V)\\
&=\H_{21}{\rm veck}(\U)+\H_{22}{\rm vec}(\V).
\end{aligned}
\end{equation}
This completes our proof.

\noindent {\bf Proof of Proposition \ref{pro4}.}
When the Riemannian Newton's vector $\bxi$ is an ascent direction of $g_{\epsilon}(\bga)$, we assert that there exists an integer $t>0$ satisfying
\begin{equation}\notag
g_{\epsilon}({{\rm Retr}_{\bga}(\sigma^t\bxi)}) \geq g_{\epsilon}({\bga})+\alpha \sigma^t\|\bxi\|^2_{\rm F}.
\end{equation}
The assertation could be proved by applying the standard argument for Armijo condition in vector spaces, see \citet[Lemma 3.1]{nocedal2006numerical}. Combining the property of the surrogate function, we then immediately obtain
\begin{equation}\notag
\cV_{n,\epsilon}^2({{\rm Retr}_{\bga}(\sigma^t\bxi)}^{\top}\Z,\Y ) \geq g_{\epsilon}({{\rm Retr}_{\bga}(\sigma^t\bxi)}) \geq g_{\epsilon}({\bga})+\alpha \sigma^t\|\bxi\|^2_{\rm F} = \cV_{n,\epsilon}^2({\bga}^{\top}\Z,\Y)+\alpha \sigma^t\|\bxi\|^2_{\rm F}.
\end{equation}
Thus, we complete the proof.

\noindent {\bf Proof of Proposition \ref{pro5}.}
Since the sequence $\left\{ \cV_{n,\epsilon}^2({\bga^{(t)}}^{\top}\Z,\Y )  \right\} $ is increasing and bounded above, $  \cV_{n,\epsilon}^2({\bga^{(t+1)}}^{\top}\Z,\Y )- \cV_{n,\epsilon}^2({\bga^{(t)}}^{\top}\Z,\Y )$ converges to $0$. According to the Proposition 4, there exists an integer $s_{t}>0$ satisfying
\begin{equation}\notag
\cV_{n,\epsilon}^2({\bga^{(t+1)}}^{\top}\Z,\Y )- \cV_{n,\epsilon}^2({\bga^{(t)}}^{\top}\Z,\Y ) \geq \alpha \sigma^{s_t}\|\bxi^{(t)}\|^2_{\rm F}.
\end{equation}
The above inequality implies that $\|\bxi^{(t)}\|_{\rm F}$ converges to zero. Recall that $\mbox{Hess}\, g_{\epsilon}(\bga^{(t)})[\bxi^{(t)}]=-\mbox{grad}\,g_{\epsilon}(\bga^{(t)})$, and we have $\mbox{grad}\,g_{\epsilon}(\bga^{(t)})$ converges to zero. Because $g_{\epsilon}(\bga)$ minorizes $\cV_{n,\epsilon}^2(\bga^{\top}\Z,\Y)$ at the point $\bga^{(t)}$, the Riemannian gradient of $\cV_{n,\epsilon}^2(\bga^{\top}\Z,\Y)$ and $g_{\epsilon}(\bga)$ are equal when evaluated at $\bga^{(t)}$. Thus, we prove the conclusion that $\mbox{grad}\,\cV_{n,\epsilon}^2({\bga^{(t)}}^{\top}\Z,\Y)$ converges to zero.

\noindent {\bf Proof of Proposition \ref{pro6}.}
Since $\cV_{n}^2(\hat{\bga}_{\epsilon}^{\top}\Z,\Y )\leq \cV_{n}^2(\hat{\bga}^{\top}\Z,\Y )$ by the definition of $\hat{\bga}$, we have
\begin{eqnarray*}
	0\leq \cV_{n}^2(\hat{\bga}^{\top}\Z,\Y )- \cV_{n}^2(\hat{\bga}_{\epsilon}^{\top}\Z,\Y ) &\leq&
	\cV_{n}^2(\hat{\bga}^{\top}\Z,\Y )- \cV_{n,\epsilon}^2(\hat{\bga}^{\top}\Z,\Y )\\
	&&+\cV_{n,\epsilon}^2(\hat{\bga}_{\epsilon}^{\top}\Z,\Y )
	-\cV_{n}^2(\hat{\bga}_{\epsilon}^{\top}\Z,\Y ) \\
	&\leq& |\cV_{n}^2(\hat{\bga}^{\top}\Z,\Y )- \cV_{n,\epsilon}^2(\hat{\bga}^{\top}\Z,\Y )|\\
	&&+|\cV_{n,\epsilon}^2(\hat{\bga}_{\epsilon}^{\top}\Z,\Y )
	-\cV_{n}^2(\hat{\bga}_{\epsilon}^{\top}\Z,\Y )| \\
\end{eqnarray*}
The right side of the above inequality goes to zero because $\cV_{n,\epsilon}^2(\bga^{\top}\Z,\Y)$ converges to $\cV_{n}^2(\bga^{\top}\Z,\Y)$ uniformly on the Stiefel manifold. Then, for a limit point $\bga^{*}$ of the sequence $\left\{ \hat{\bga}_{\epsilon_m} \right\}_{m\geq 1}$ with $\epsilon_m \downarrow 0$, we have
\begin{equation}\notag
\underset{m\to\infty}{\rm lim } \, \cV^2_{n}(\hat{\bga}_{\epsilon_m}^{\top}\Z,\Y)= \cV^2_{n}({\bga^{*}}^{\top}\Z ,\Y)=\cV^2_{n}(\hat{\bga}^{\top}\Z ,\Y)= \underset{ \bga \in \St(p,d) }{\rm max } \cV^2_{n}(\bga^{\top}\Z ,\Y).
\end{equation}
by the continuity of $\cV^2_{n}(\bga^{\top}\Z ,\Y)$. Thus, we complete our proof.

\noindent {\bf Proof of Theorem \ref{thm1}.}
For the first part, we have
\begin{equation}\notag
\|\hat{\bga}_{\epsilon}^{(t)}-\hat{\bga}\|_{\rm F}\leq \|\hat{\bga}_{\epsilon}^{(t)}-\hat{\bga}_{\epsilon}\|_{\rm F}+\|\hat{\bga}_{\epsilon}-\hat{\bga}\|_{\rm F},
\end{equation}
where $\hat{\bga}_{\epsilon}$ is a maximizer of $\cV_{n,\epsilon}^2(\bga^{\top}\Z,\Y) $ over the Stiefel manifold, and $\hat{\bga}$ is a limit point of $\left\{ \widehat{\bga_{\epsilon}} \right\}_{\epsilon\geq 0}$ as $\epsilon \downarrow 0$. By Proposition 5, we know the first term becomes arbitrarily small for sufficiently large $t$, whereas the second term does so for sufficiently small $\epsilon$ by Proposition 6. The limit point $\hat{\bga}$ is a maximizer of $\cV_{n}^2(\bga^{\top}\Z,\Y)$ over the Stiefel manifold by Proposition 6. For the second part, we have
\begin{eqnarray*}
	|\cV_{n,\epsilon}^2(  \hat{\bga}_{\epsilon}^{(t)\top} \Z ,\Y)-\cV_{n}^2(\hat{\bga}^{\top}\Z ,\Y )| &\leq&
	|\cV_{n,\epsilon}^2(  \hat{\bga}_{\epsilon}^{(t)\top} \Z ,\Y)-\cV_{n,\epsilon}^2(\hat{\bga}_{\epsilon}^{\top}\Z ,\Y )|\\
	&& +|\cV_{n,\epsilon}^2(\hat{\bga}_{\epsilon}^{\top}\Z ,\Y )- \cV_{n}^2(\hat{\bga}_{\epsilon}^{\top}\Z ,\Y )| \\
	&&+ |\cV_{n}^2(\hat{\bga}_{\epsilon}^{\top}\Z ,\Y ) -\cV_{n}^2(\hat{\bga}^{\top}\Z ,\Y )| .
\end{eqnarray*}
The first and third term in the right-hand side vanish respectively by the continuity of $\cV^2_{n,\epsilon}(\bga^{\top}\Z,\Y)$ and $\cV^2_{n}(\bga^{\top}\Z,\Y)$; the second term by the uniform convergence of $\cV^2_{n,\epsilon}(\bga^{\top}\Z,\Y)$ to $\cV^2_{n}(\bga^{\top}\Z,\Y)$, as shown in the proof of Proposition 1. Thus, we have completed our proof.

\bibliographystyle{cas-model2-names}

\bibliography{References}

\end{document}